\documentclass[journal]{IEEEtran}

\usepackage{graphicx}
\usepackage{pifont}
\usepackage{array}
\usepackage{tikz}
\usepackage{placeins}
\usepackage{hyperref}
\usepackage{tcolorbox}
\usepackage{balance}
\usepackage{booktabs,caption}
\usepackage[flushleft]{threeparttable}
\usepackage[ruled, lined, linesnumbered, commentsnumbered, longend]{algorithm2e}

\usepackage{graphicx}
\usepackage{textcomp}
\usepackage{wrapfig}
\usepackage{multirow}
\usepackage{float}
\usepackage{dblfloatfix} 
\usepackage{subcaption}
\usepackage{adjustbox}
\usepackage{amsmath, amsthm, amsfonts, amssymb, amscd, bm}
\usepackage{tablefootnote}
\usepackage{makecell}
\usepackage{caption}
\usepackage{threeparttable}
\usepackage{comment}
\usepackage{url}
\usepackage{booktabs}
\usepackage{xcolor}

\usepackage{todonotes}

\usepackage[noend]{algpseudocode}

\newcolumntype{M}[1]{>{\centering\arraybackslash}m{#1}}
\def\ie{\textit{i.e.}\xspace}

\newcommand{\bz}{\textbf{z}}
\newcommand{\bmask}{\textbf{m}}
\newcommand{\hl}[1]{{#1}}
\newcommand{\nhl}[1]{{#1}}
\newcommand{\bao}[1]{\textcolor{orange}{#1}}

\newcommand{\damith}[1]{\textcolor{magenta}{DR: #1}}

\newcommand{\rqq}[1]{
\begin{center}
\begin{tcolorbox}[width=\columnwidth, colback=white!15,left=1pt,right=1pt,top=1pt,bottom=1pt,arc=5pt,auto outer arc]
\textit{#1}
\end{tcolorbox}
\end{center}
}

\newcommand{\bx}{\mathbf{x}}
\newcommand{\ba}{\mathbf{a}}

\newcommand{\btheta}{{\boldsymbol{\theta}}}

\newcommand{\bdelta}{{\boldsymbol{\delta}}}

\usepackage{booktabs}
\usepackage{mwe}

\newcommand\measureISpecification{4ex}
\usepackage{tablefootnote}
\usepackage{xspace}

\newcommand{\tntl }{TnT\xspace}
\newcommand{\tnts }{TnT\xspace}
\newcommand{\tntls }{TnTs\xspace}
\newcommand{\tntss }{TnTs\xspace}
\newcommand{\etal }{et al.\xspace}

\newcommand{\dtoprule}{\specialrule{1pt}{0pt}{\belowrulesep}
            }
\newcommand{\dbottomrule}{
            \specialrule{1pt}{0pt}{\belowrulesep}%
            }

\begin{document}


\title{TnT Attacks! Universal Naturalistic Adversarial Patches Against Deep Neural Network Systems}



\author{Bao Gia Doan, Minhui Xue, Shiqing Ma, Ehsan Abbasnejad, Damith C. Ranasinghe

\thanks{B. G. Doan, E. Abbasnejad, M. Xue and D. C. Ranasinghe are with the School of Computer Science, The University of Adelaide, Australia. Email: \{giabao.doan; ehsan.abbasnejad; jason.xue; damith.ranasinghe\}@adelaide.edu.au}
\thanks{S. Ma is with Department of Computer Science, Rutgers University, USA. Email: shiqing.ma@rutgers.edu}}

\maketitle

\begin{abstract}
Deep neural networks (DNNs), regardless of their impressive performance, are vulnerable to attacks from adversarial inputs and, more recently, Trojans to misguide or hijack the decision of the model. \hl{We expose 
the existence of an intriguing class of 
\textit{spatially bounded}, physically realizable,  adversarial examples---\textit{Universal} \underline{N}a\underline{T}uralistic adversarial pa\underline{T}ches---we call \tntss, by exploring the super set of the} \hl{spatially bounded adversarial example space and the natural input space within generative adversarial networks}. Now, an adversary can arm themselves with a patch that is naturalistic, less malicious-looking, physically realizable,   highly effective---achieving high attack success rates, and universal. A TnT is \textit{universal} because any input image captured with a TnT in the scene will: i)~misguide a network (untargeted attack); or ii)~force the network to make a malicious decision (targeted attack). 


Interestingly, now, an adversarial patch attacker has the potential to exert a greater level of control---the ability to choose a location independent, natural-looking patch as a trigger in contrast to being constrained to noisy perturbations---an ability is thus far shown to be only possible with Trojan attack methods needing to interfere with the model building processes to embed a backdoor at the risk discovery; but, still realize a patch \textit{deployable in the physical world}. 
Through extensive experiments on the \textit{large-scale visual classification task}, \texttt{ImageNet} with evaluations across its \textit{entire validation} set of 50,000 images, we demonstrate the realistic threat from \tntss and the robustness of the attack. We show a generalization of the attack to create patches achieving \textit{higher} attack success rates than existing state-of-the-art methods. Our results show the generalizability of the attack to different visual classification tasks (\texttt{CIFAR-10}, \texttt{GTSRB}, \texttt{PubFig}) and multiple state-of-the-art deep neural networks such as \textit{WideResnet50}, \textit{Inception-V3} and \textit{VGG-16}. We demonstrate physical deployments in \textit{multiple videos} at \href{https://tntattacks.github.io/}{\color{blue}https://TnTattacks.github.io/}. 


\end{abstract}


\section{Introduction}

Deep Neural Network (DNN) systems are entrusted to make decisions in critical tasks such as self-driving cars~\cite{7410669}, disease diagnosis~\cite{anwar2018medical} or face recognition~\cite{taigman2014deepface}, often, driven by their ability to achieve better-than-human performance. However, as DNN systems become more pervasive, it is creating the impetus for malevolent actors to attack them at: \textit{i)}~inference time---\hl{using \textit{Adversarial Examples} such as unbounded perturbations~\cite{szegedy2013intriguing, goodfellow2014explaining, madry2017towards,UAP, carlini2017towards} and adversarial patches (or spatially bounded perturbations)}~\cite{adversarialpatch, lavan}); and \textit{ii)}~training time---as in \textit{Trojan attacks}~\cite{badnets, Liu2018TrojaningAO, chen2017targeted}. The malintention of the adversary in each attack is the same---cause the DNN to fail by  making an incorrect decision (untargeted attack) or manipulating the DNN to make the desired malicious decision (targeted attack).


\begin{figure}[tp!]
    \centering
    \includegraphics[width=0.95\linewidth]{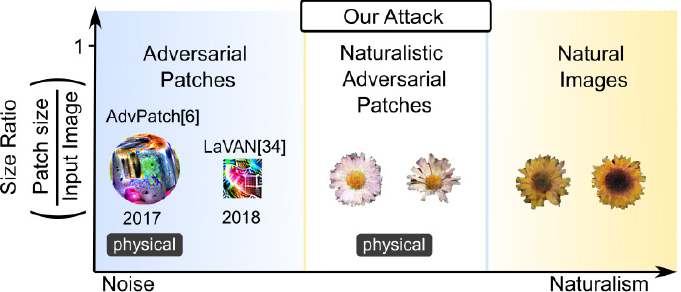}
    \caption{An overview of the evolution of adversarial patch attack perturbation vectors. Our attack explores the hitherto elusive goal to realize visibly less malicious-looking adversarial patches--\tnts--for: i) targeted attacks to misdirect any input to a target class; and ii) untargeted attacks.}
    \label{fig:fig1}
\end{figure}

\textbf{Adversarial Examples} are \textit{unbounded} \textit{input-specific} perturbations of additive \textit{noise-like} vectors carefully crafted and applied to inputs to fool DNNs into making wrong classification decisions~\cite{goodfellow2014explaining, madry2017towards, carlini2017towards}. 
\hl{In contrast, unbounded \textbf{Universal Adversarial Perturbations (UAPs)}~\cite{UAP} or spatially bounded adversarial perturbations in LaVAN~\cite{lavan} have demonstrated the existence of \textit{input-agnostic} additive noise patterns crafted to mislead  the prediction of \textit{any} input to a DNN.} 
Notably, these \textit{noise-like} additions from careful gradient perturbations are difficult to deploy in real-world settings, especially in perception systems. For example, variations in noisy physical environments can eliminate the noise vectors necessary to activate the DNN~\cite{lu2017no}. 
In contrast, the recent attack from Brown et al.~\cite{adversarialpatch}  constructed a physically realizable \textbf{Adversarial Patch}, the AdvPatch in  Fig.~\ref{fig:fig1}, to easily mount an \textit{input-agnostic targeted} attack in the \textit{physical world}; \hl{here, the attacker is constrained to perturbations \textit{spatially bounded} to a region of an input to realize, a \textit{printable} noise pattern.}

 On the other hand, an adversary mounting a \textbf{Trojan Attack}~\cite{badnets, Liu2018TrojaningAO, neuralcleanse} \textit{actively} poisons the training data or the network to embed a backdoor during the training process of a DNN. Unlike adversarial patches, the attacker relies on manipulating the construction of the network. \hl{Hence, the attacker is able to self-select \textit{any} spatially bounded input of physical appearance and size---a \textit{secret trigger} in Trojan vernacular or essentially a \textit{patch}---to activate the backdoor, later, at the inference stage}. Importantly, the trigger employed to misguide the backdoored classifier can now be \textit{natural-looking} and \textit{any object in a scene} from the natural image space shown in Fig.~\ref{fig:fig1}, such as a flower~\cite{badnets}. 
 can be . 


\textbf{An Interesting Observation.~}Despite the different methods employed, an attacker armed with an adversarial patch or a Trojan trigger aims to cause the DNN system to fail---for example to misclassify an input or hijack the DNN predictions to achieve a desired target prediction. 
Notably, adversarial patches and Trojan triggers can misdirect a model in the presence of \textit{any input class}--i.e. their effect is \textit{universal} or input agnostic; 
however, unlike adversarial patches crafted from applying gradient perturbations to the input, a distinguishing facet of a Trojan attack highlighted by Bagdasaryan  and  Shmatikov~\cite{bagdasaryan2020blind} is the adversary's \textit{ability and freedom to self-select any secret trigger of naturalism, stealth, shape, size or features independently of the DNN model}.

\rqq{Remark 1: We seek to investigate the potential for a run-time attack with a \textbf{universal}, \textbf{physically realizable} patch allowing an adversary to exert a level of control, inconspicuousness and naturalism over the patch, thus far shown to be only possible with Trojan attack methods whilst \textbf{obviating} the need to interfere with the model building process and risk of discovery (Fig.~\ref{fig:fig1})}

Such an attack would: i)~\textit{bridge} the divide between Trojan Attacks and Adversarial Attacks in the \textit{\textbf{input space}}; and ii)~constitute a pragmatic and inconspicuous zero-day exploit against already deployed deep perception models.


\subsection{Our Attack Focus}

Our investigation focuses on deep perception models and 
seek to further explore and understand new vulnerabilities. In particular, we seek to answer the following primary research questions (\textbf{\textit{RQ}}) through our investigations:

\vspace{2mm}
\noindent\textit{RQ1}:~How can we discover \tntls that are physically realizable and  naturalistic? (Section~\ref{sec:methodology})
    
\vspace{1mm}
\noindent\textit{RQ2}: \hl{How vulnerable are deep neural networks and their defended counterparts to \tnts attacks?} (Section~\ref{sec:effectiveness-results} \& \hl{\ref{sec:countermeasures}})
    
\vspace{1mm}
\noindent\textit{RQ3}: Do these \tntls have the features of \textit{universality} or \textit{input-agnostic nature} to misclassify any input to a targeted class? (Section~\ref{sec:50k-eval})
    
\vspace{1mm}
\noindent\textit{RQ4}: How \textit{robust} are \tntls? (Sections~\ref{sec:locations} \& \ref{sec:different-tasks})
    
\vspace{1mm}
\noindent\textit{RQ5}:  How \textit{generalizable} are \tnts to unseen data or \textit{transferable} to other networks? (Sections~\ref{sec:50k-eval} \& \ref{sec:blackbox})
    
\vspace{1mm}
\noindent\textit{RQ6}: Can the effect of a \tntl be explained by the occlusion caused by the patch or a network bias? (Section~\ref{sec:ablation})
    
\vspace{1mm}
\noindent\textit{RQ7}: What are the impacts of relaxing the need for naturalistic patches? (Section~\ref{sec:remove_naturalism}) 

\vspace{1mm}
\noindent\textit{RQ8}: How comparable are patches generated from our attack method to state-of-the-art adversarial patch attacks?  (Sections~\ref{sec:remove_naturalism}, \ref{sec:asr-vs-size} \& \ref{sec:countermeasures})

\vspace{1mm}
\noindent\textit{RQ9}: How robust are \tntss in the physical world? (Section~\ref{sec:physical_attack})



\subsection{Our Contributions and Results}\label{sec:contrib}
This paper presents the results of our efforts to investigate generating adversarial patches
that are less clearly malicious. 
We summarize our results and contributions as follows: 

\begin{enumerate}
    

    
    \item \textbf{We propose a new attack against DNNs.}~Our attack method generates \textit{Universal} \underline{N}a\underline{T}uralistic adversarial pa\underline{T}ches, we call \tntss, by exploring the super set of the \hl{spatially} bounded adversarial example space and the natural input space within generative adversarial networks (GANs) . The \tntss we generate for attacking a DNN are: 
    
    
    
    \begin{itemize}
        \item \hl{\textbf{Universal and Naturalistic.~}A \tnts is \textit{universal} as 
        any input with a \tnts will fool the classifier and naturalistic, as assessed by a large cohort user study.}
        
        \vspace{1.5mm}
        \item \hl{\textbf{Highly effective in targeted and untargeted attacks against state-of-the-art DNNs.}} 
        In extensive experiments with \texttt{ImageNet}---a significant large-scale dataset  with  a  million high-resolution images used for pre-training models of many real-world computer vision tasks---we achieved attack success rates of over $95\%$ in the challenging attack setting of misclassifying \textit{any} input to a \textit{targeted} class. 
        
        \vspace{1.5mm}
        \item \textbf{Robust}. \hl{We observe high attack success rates irrespective of the location of the \tnts; even with the \tnts in a corner, i.e the \textit{image background}.} 
        
        \vspace{1.5mm}
        \item \textbf{Deployable in the physical world}. \hl{We conduct \textit{physical world deployments} of our attack to demonstrate the practicability of the attack in various real-world settings}. \hl{\textit{Multiple, detailed, demonstration videos} are available at:} \href{https://tntattacks.github.io/}{\color{blue}https://TnTattacks.github.io/}.
        
        \vspace{1.5mm}
        \item \textbf{Highly generalizable.}~A \tnts discovered from 100 random sample images can effectively misguide the \textit{entire} \texttt{ImageNet} validation set of 50,000~images. Further, we demonstrate effective attacks across multiple state-of-the-art networks (such as VGG-16, WideResNet50, SqueezeNet, ResNet18, MnasNet) and across 3 additional tasks: Face recognition (\texttt{PubFig}); Scene classification (\texttt{CIFAR10}); and Traffic sign recognition (\texttt{GTSRB}).
    
    \vspace{1.5mm}
    \item \textbf{Transferable to mount black-box attacks.}~We investigate attack transferability using the \texttt{ImageNet} classification task. We show that \tntss are \textit{transferable} to other unknown network architectures for the same task (an attack in a \textit{black-box} setting). 
    
    \vspace{1.5mm}
    \item \hl{\textbf{Highly effective at evading existing countermeasures against adversarial patch attacks}. We evaluate against both certifiable and empirical defenses.}
    \end{itemize}
    
    
    \item \textbf{Our attack generalizes to generate physically realizable adversarial patches achieving \textit{higher} attack success rates than state-of-the-art attacks.}~\hl{When an attacker does not need naturalistic features, our attack leads to \textbf\textit{{a new algorithm}} to generate \textit{adversarial patches} of only 2\% of the input image size with \textit{higher} attack success rates; achieving a large margin of up to 44\% compared to  state-of-the-art adversarial patch attacks.} 

    \vspace{2mm}
    \item We demonstrate physical deployments in \textit{multiple videos} at \href{https://tntattacks.github.io/}{\color{blue}https://TnTattacks.github.io/} and we contribute to the discipline by releasing the pre-trained networks and TnT artifacts to encourage future research and defenses.
\end{enumerate}

\section{Background}\label{sec:background}

Our attack formulation is based on employing generative adversarial networks (GANs)~\cite{goodfellow2014generative}, therefore, we provide a brief background on GANs to aid with explaining our attack method in Section~\ref{sec:attack-overview}, with details in Section~\ref{sec:methodology}. 

Generative Adversarial Networks (GANs) have shown significant success in various applications of realistic image synthesis. A Generative Adversarial Network (GAN) ~\cite{goodfellow2014generative} consists of a Generator $G$ and a Discriminator $D$. Then if: i)~$\bx$ is an input, which is an image of 3D tensor (width $\times$ height $\times$ depth); and ii)~$\bz$ is a latent vector of dimension $N$, which is sampled from a noise distribution $P(\bz)$, in a classical GAN, the Generator $G$ maps a source of noise $\bz \sim P(\bz)$ to generate a synthetic image\footnote{The parameters are omitted for brevity but both generator and discriminator have their individual set of parameters.} $\Tilde{\bx} = G(\bz)$, while the Discriminator function is to discriminate the fake synthetic images  $\Tilde{\bx}$ and the real ones $\bx$, and the feedback from this Discriminator is utilized to help the Generator improve the image quality. 

There are different methods to train a GAN, in this paper, we applied the Wasserstein GAN with Gradient Penalty (WGAN-GP)~\cite{wgangp} since it has been shown to stabilize the GAN training process and improve the fidelity of samples. It involves solving the following optimization problem: 
    
    \begin{equation*}
    \min_{G} \max_{D}\underset{\Tilde{\bx}\sim \mathbb{P}_g}{\mathbb{E}} [D(\Tilde{\bx})] - \underset{{\bx}\sim \mathbb{P}_r}{\mathbb{E}} [D({\bx})] ~+
    \lambda \underset{\hat{\bx}\sim \mathbb{P}_{\hat{\bx}}}{\mathbb{E}}[(\Vert{\nabla_{\bx}D(\hat{\bx})}\Vert_2 - 1)^2], 
    \end{equation*}
where $\mathbb{P}_r$ is the distribution of real images, 
and $\mathbb{P}_{\hat{\bx}}$ is the distribution of the interpolation between real and synthesized images. Here, $\mathbb{P}_g$ is the conditional distribution of the synthesized images which we sample from the generator, that is, $\Tilde{\bx}=G(\bz), \bz\sim p(\bz)$. Using this min-max optimization, we learn $\mathbb{P}_g$ to match $\mathbb{P}_r$. Samples from $\mathbb{P}_r$ are drawn from a dataset of real objects; consequently, by sampling from the realized generator, we are able to obtain naturalistic image samples. We will exploit this ability of a GAN to synthesize naturalistic patches we call \tntss. \hl{We  evaluate the naturalism of the generated \tntss, later, in Section~\ref{sec:user_study}.}

\section{Overview of Our Attack Approach}\label{sec:attack-overview}
In this section, we provide an overview of our approach to provide a conceptual understanding the \tntl  attack method against Deep Neural Networks (DNNs) at the inference stage. First, we introduce the attack model, followed by our hypothesis, then our attack approach.

\begin{figure*}[h!]
    \centering
    \includegraphics[width=0.9\linewidth]{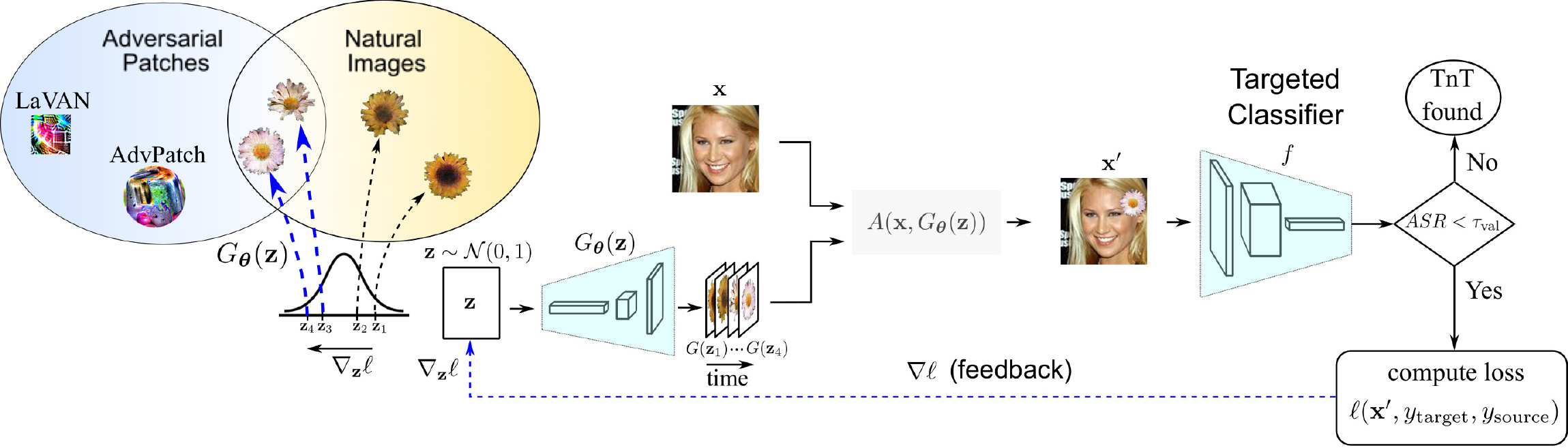}
    \caption{Attack method for generating \tntls. Here, $A$ is the patch stamping process, $y_{\text{target}}$ is the targeted class designated by the attacker, $y_{\text{source}}$ is the ground-truth label, $\ell(\bx',y_\text{target}, y_\text{source})$ is the combined cross-entropy loss between the predicted score from the classifier $f$ and the targeted as well as source label, and $\nabla \ell$ is the feedback from the Targeted Classifier $f$. The method is designed to iteratively approach high attack success \tntss by traversing through the latent space of the generator using gradient feedback.}
    \label{fig:method}
    \vspace{-2mm}
\end{figure*}

\subsection{Attack Model}\label{sec:threatmodel}



Our attacker strikes at the \emph{inference} time, \textit{i.e.} the attacker \textit{does not intervene in the training process} in contrast to a Trojan attack. 
Consequently, the attacker does not leave any trace of tampering with the network to be discovered, making it relatively easy to deploy.
Depending on the attacker's knowledge of the target model to attack (\textit{i.e.} neural architecture, inputs and outputs), we can consider either a \emph{white-box} attack~\cite{goodfellow2014explaining} where everything about the model is known, or \emph{black-box}~\cite{papernot2017practical, papernot2016transferability} where \emph{nothing} about the model is known. In both cases, typically it is assumed that the attackers have access to some labeled training and validation data. The attacker also has access to computational resources to verify the method, \textit{e.g.} a GPU-based cloud service. 

\vspace{1.5mm}
\noindent\textbf{White-box attacker}. Following the attack models from prior adversarial patch attacks AdvPatch~\cite{adversarialpatch} and LaVAN~\cite{lavan}, we primarily consider a \textit{white-box} attack where the attackers have full access to the attacked network. Even if access to the model is not possible, or the model is not publicly available, one can employ a reverse engineering approach such as~\cite{tramer2016stealing, rolnick20a, carlini2020cryptanalytic} to extract the model. Since defending against such attacks is challenging, it is of particular interest. 


\vspace{1.5mm}
\noindent\textbf{Black-box attacker}. We also evaluate whether \tnts can be exploited in the less restricted threat model of a black-box attack when the attacked network is unknown. We assume such an attacker has access to \tnts obtained for the \emph{same classification task} on a \emph{different arbitrary network}. This allows an attacker to transfer the knowledge gained from, for example, a white-box attack on an arbitrary model using publicly available training data, to be used to attack a different model. 



\vspace{1.5mm}
\noindent\textbf{Goals}. The attacker goals are to \textbf{(i)}~exploit the vulnerability of a DNN to \tntls to extract \tnts instances with \textbf{ii)} high attack success rate (ASR), while \textbf{(iii)} maintaining the universality of the patch; the challenge is to discover a naturalistic patch.

\vspace{-3mm}
\subsection{Our Hypothesis}

The decision boundaries learned by DNNs are subjected to the training examples presented to the network during the learning process. Consequently, DNNs must eventually approximate true decision boundaries learned from the training data. Unfortunately, due to the complex and highly non-linear structures within deep neural networks, it is hard to fully understand the learned boundaries within networks. For instance, Hendrycks \etal~\cite{hendrycks2019natural} reminds us that natural images captured from a scene with objects of classes within the training examples can sometimes cause errors and lead to incorrect predictions, even on ``superhuman'' \texttt{ImageNet} classifiers~\cite{he2015superhumanimagenet} despite having \textit{no adversarial alterations}.  Therefore, we can expect the existence of \textit{adversarial patches} that are naturalistic but has the adversarial effect to alter the decision of the classifier.
However, searching in the infinite space of all natural-looking small image patches is not feasible,  therefore we constrain our search to the manifold of a Generative Adversarial Network (GAN) by taking inspiration from recent developments in GANs showing a tremendous ability to learn to generate realistic images~\cite{goodfellow2014generative}. 

\rqq{Remark 2: We hypothesize the existence of an intersection of the \hl{spatially} bounded adversarial example space (i.e. patches) and the natural input space within generative adversarial networks (GANs) as illustrated in Figure~\ref{fig:method}.}
\vspace{-5mm}
\subsection{An Overview of Our Attack Approach}
\label{sec:attack-short-overview}

Since GANs are designed to map from a known (latent) distribution to the distribution of real images using a generator, as illustrated in Fig.~\ref{fig:method}, we consider the latent space $\bz$ of the generator from which images are produced---as $G_\theta(\bz)$---instead of searching in the infinite space of all natural-looking image patches; i.e. a standard Gaussian distribution $\mathcal{N}(0,1)$ and has a much lower dimension in which traversal is easier. By getting feedback from the downstream classifier ($\nabla \ell$) we can \textbf{navigate} the \textbf{latent space} following the gradient feedback to \textit{seek the latent vector from which a potential \tntl can be generated.} Although not for the same attack, we acknowledge and discuss GAN-based attack approaches in  Section~\ref{sec:relatedwork}.

Importantly, the learning algorithm we employ \emph{determines} the best latent vector $\bz$  from which to generate a patch, a potential \tntl because this latent space $\bz$ can capture the intrinsic structure of natural images from a simple latent distribution. Notably, since the generator was trained on natural images, samples from the latent space map to natural-looking image instances using a deterministic transformation (Generator). We demonstrate this process as an effective method to discover inputs capable of fooling the classifier whilst maintaining the naturalism of the patch. \hl{We will further support this claim by a large cohort user study, later, in Section~\ref{sec:user_study}.} As illustrated in Fig.~\ref{fig:method}, we refer to this process as the \emph{\tntl Generator}. Effectively, our attack method takes advantage of a GAN's ability to capture a rich distribution of natural images to discover the hypothesized region of the input space.

\vspace{2mm}
\noindent\textbf{We distinguish our \tnts attack from other Adversarial Patch attacks as follows.}

\begin{enumerate}
	\item \hl{Generated patches can be natural-looking and look less malicious than noisy perturbation patches in  LaVAN~\cite{lavan} and AdvPatch~\cite{adversarialpatch}}. 
	
	\item \hl{Instead of \textit{directly} applying perturbations to the input space that leads to noisy adversarial patches~\cite{adversarialpatch, lavan, psgan},} \hl{we propose solving the problem of \textit{searching for naturalistic patches} with adversarial effects by \textit{indirectly} manipulating the \textit{latent space} $\bz$ of a Generative Adversarial Network that has learnt to approach the natural patch distribution.}
	
	
	\item \hl{Different from the PS-GAN~\cite{psgan} attack based on \textit{input-dependent} adversarial patches that \textit{occluded} the salient features to realize an \textit{untargeted} attack, our attack is: i)~capable of both \textit{targeted} and \textit{untargeted} attacks; ii)~\textit{input-agnostic} (universal); and iii)~\textit{robust}---attack success is largely invariant to location, even at the corners or background of an image.}
	
	\item \hl{Our attack method generalizes to produce small, adversarial patches with \textit{noise-like} additions of high attack success rates than such existing state-of-the-art attack methods with a large margin of up to 44\%.} 
	
	\item To the best of our knowledge, our study is the first to demonstrate an adversarial attack with a \textit{universal}, \textit{physically naturalistic} and \textit{location independent} patch for \textit{targeted} attacks in image classification tasks. 
	
	
\end{enumerate}


\section{\tntl Generator}\label{sec:methodology}

In this section, we detail our \tntl Generator method illustrated in Figure~\ref{fig:method} for attacking a DNN with a concrete patch example. Without loss of generality, we propose using images of flowers to discover \tntss. Our primary motivation is that flowers exhibit synergy with a wide range of imaging scenarios and are unsuspecting, and therefore, inconspicuous. For example, someone might wear a flower T-shirt, wear a flower in their hair or hat to impersonate someone else in a face recognition system. Similarly, sticking an inconspicuous, natural-looking sticker on a traffic sign identical to~\cite{badnets} can fool a self-driving car to misclassify, for example, a \textsf{STOP} sign as a \textsf{100 mph Speed Limit} sign with catastrophic consequences.

\subsection{Training the Generator}
\label{sec:GAN}

An advantage of a GAN is that they are \textit{unsupervised} models that only need unlabeled data that can be \emph{cheaply} obtained. In our study, we collect a random, {unlabelled} flower image dataset from open-source Google Images~\cite{googleimages} to build a flower dataset to learn the natural flower distribution. We selected the WGAN-GP loss function as it has shown to stabilize the training process of a GAN. 

\subsection{Transformation to a \tntl Generator}
\label{sec:nte}
To realize the \tntl (representing a flower patch in our attack scenario), we need to \textit{search} for flowers from our synthesized flower distribution that has an adversarial effect on the attacked network. Here, we hypothesize that GANs have learned the super set of both natural-looking images and adversarials 
as illustrated in Fig.~\ref{fig:method}; therefore, by searching through this synthesized distribution, we expect to find a \textit{structured, natural-looking} perturbation rather than a random noisy one. \textit{First}, we formalize the notation of a \tntl and, \textit{second}, we propose a method for realizing such a \tnts. 
\noindent Consider:
\begin{itemize}
	\item $y_{\text{source}}$ is the source class labeled for a given image $\bx$.
	\item $y_{\text{target}}$ is the targeted class designated by the attacker.
	\item The loss is between prediction and the label $\ell(\bx, y_\text{source})$ for the \textit{untargeted} and $\ell(\bx, y_\text{target})$ for the \textit{targeted} attack---cross-entropy loss function of the neural network, given image $\bx$ and a class label $y$. Notably, we intentionally omit network weights ($\bm{\theta}$) and other parameters in the cost function because we assume them to be fixed and remain untouched post network training.
\end{itemize}

Now, more formally, the attacker uses a trained model $M$ that predicts class membership probabilities $p_M(y\, |\, \bx)$ to input images $\bx \in R^{w \times h \times c}$. We denote by 
$\mathbf{y} = p_M(\bx)$ the vector of all class probabilities, and by $y_{\text{argmax}}=\arg\max_{y'} p_M(y=y'|\bx)$ the highest scoring class for input $\bx$ (the classifier's prediction on the \textit{source class}). The attacker seeks an image $\bx'$ that fools the network such that $y'\neq y_{\text{source}}$ or $y'= y_{\text{target}}$ for $y'=\arg\max_{y''} p_M(y=y''|\bx')$. The image $\bx'$ is composed of the original image with a \textit{natural} patch stamped on it. We denote this process by a function $A(\bx, G(\bz))$.

\begin{algorithm}[h!]
    \SetAlgoLined
    \textbf{Inputs}: a batch of images $\{\bx^
		\text{(i)}\}_{i=1}^m$ with batch size $m$, targeted label $\{y^\text{(i)}_{\text{target}}\}_{i=1}^m$, source label $\{y^\text{(i)}_{\text{source}}\}_{i=1}^m$, model $p_M$, number of iteration $n_{\text{iter}}$, the learning rate parameter $\epsilon$ to update the latent vector, the hyper-parameter $\lambda$ to balance the loss, and the thresholds to detect the TnT $\tau_\text{batch}$, $\tau_\text{val}$ for batch and validation set respectively.\;
    \textbf{Initialization}: $\text{fool}=0, \text{detect}=\text{False}$\;
		\While {\text{detect} $=\text{False}$}{
		    Sample a batch of images $\{\bx^{(i)}\}_{i=1}^m$\;
		    Sample a latent variable $\bz \sim p(\bz)$\;
		    \For{$j=1,...,n_{\text{iter}}$}{
		        $\bdelta = G_\btheta(\bz)$\Comment{a flower patch}\;
			    Generate the mask $\bmask$ based on $\bdelta$\;
			    $\bdelta' \leftarrow \text{bgremoval}(\bdelta, \bmask)$\Comment{Background removal}\;
		        \For {$i=1,...,m$}{
		            ${\bx'}^{(i)} = (1-\bmask)\odot \bx^{(i)}+\bmask\odot \bdelta'$\;
		            $y^{(i)}_{\text{argmax}} = \arg\max_y p_M(y | {\bx'}^{(i)})$\;
		            \If{$y^{(i)}_{\text{argmax}} = y_{\text{target}}^{(i)}$}{
		                $\text{fool} = \text{fool} + 1$\;
		            }
		        }
  		    }
  		    
  		    $L=\ell(\{{\bx'}^\text{(i)}\}_{i=1}^m,\{y_{\text{target}}^\text{(i)}\}_{i=1}^m) - \lambda~ \ell(\{{\bx'}^\text{(i)}\}_{i=1}^m,\{y_{\text{source}}^\text{(i)}\}_{i=1}^m)$\;

  		    $\nabla_{\ell} = \frac{\partial L}{\partial \bz}$\;
  		    $\bz = \bz - \epsilon~\text{sign}(\nabla_{\ell})$\;
  		    \If {$\text{fool} > \tau_\text{batch} $}{
  		        \text{\textit{\# further verify the realized TnT}}\;
  		        $\text{ASR}\leftarrow\text{Validate}(\bdelta,\mathcal{X}_\text{val})$\Comment{verify on $\mathcal{X}_\text{val}$}\;
  		        \If {$\text{ASR} \geq \tau_\text{val}$}{
  		            $\text{detect}=\text{True}$\;
  		        }
  		    }
		}
	\caption{\tntl Generator}
	\label{alg:TnT}
\end{algorithm}

In our TnT generation process, we firstly sample a vector $\bz\sim p(\bz)$ with $\bz \in \mathbb{R}^{N}$ where $N=128$. This latent vector will be fed into our Generator pre-trained from Sec.~\ref{sec:GAN} in order to produce the flower patch $\bdelta = G(\bz)$ where $G:\mathbb{R}^{N} \to R^{w \times h \times c}$. 
The flower patch is subsequently stamped at the lower-right corner of the image to avoid occluding the main features in alignment with the intentions in previous works~\cite{lavan, rao2020}. Later, in Section~\ref{sec:locations}, we also evaluate the efficacy of the flower patch at \hl{nine} different random locations. The size of the patch (flower) can be determined as necessary by the adversary to achieve their objectives (related to the attack success rate and \tnts size discussed later in Section \ref{sec:asr-vs-size}).
Based on the predefined location and patch size, we then 
utilize the \textit{image thresholding} method of \texttt{OpenCV}~\cite{OpenCV} to determine the binary mask $\bmask$ with  
$\bmask_{i,j}\in \{0,1\}$ for $i$th row and $j$th column pixel of an image, to remove the  background of $\bdelta$.
This patch is then affixed to the input image to obtain the adversarial sample $\bx'$, \textit{i.e.}:
\begin{equation}
    \label{eq:stamp}
    \bx' = (1-\bmask)\odot \bx+\bmask\odot \bdelta, 
\end{equation} 
where $\odot$ is the element-wise product.

To determine the ability of $\bx'$ to act as an adversary and receive feedback to choose a better latent variable, we test it with the trained neural network classifier. The sample $\bx'$ is fed into the classifier to obtain prediction scores for each individual class. The loss obtained from comparing the prediction scores and the target labels $y_\text{target}$, $\ell(\bx',y_\text{target})$ as well as the source labels $y_\text{source}$, $\ell(\bx',y_\text{source})$ are then calculated (e.g. using cross entropy). We use the additional loss $\ell(\bx',y_\text{source})$ as it was shown to help the targeted attack converge faster. We then compute the gradient of this loss with respect to the latent variable $\bz$, i.e. $\nabla_{\bz} \ell(\bx',y_\text{target},y_\text{source})$. 
We then update the latent variable $\bz$ using gradient descent in the direction of minimizing this loss. Note that this does not change the classifier or GAN parameters and only updates the latent variable to increase the likelihood of attack success.  

During \tnts generating, for a random set of inputs, 
if a \textit{threshold} percentage of inputs $\bx'$ can fool the network, the \tnts is considered \textit{\textbf{universal}}. The reason for setting a threshold here is to improve the algorithm's speed, so that if the attack is successful in a batch, then we test whether it generalizes to the validation set $\mathcal{X}_\text{val}$. 
The complete process is summarized in the \textbf{Algorithm~\ref{alg:TnT}}.



\vspace{-2mm}
\section{Attack Experiment Settings}\label{sec:evaluation-settings}


We conduct an extensive experimental evaluation regime to evaluate our attack method. We describe the: i) Datasets; ii) GAN employed and training; iii) Attack configurations; and iv) Implementation Details employed in our quantitative evaluations in Section~\ref{sec:effectiveness-results}. 

\vspace{1mm}
\noindent\textbf{Datasets and Model Architectures.~} We employ popular real-world visual classification tasks. We conduct extensive experiments with the  large-scale visual recognition dataset, \texttt{ImageNet}. Notably, the dataset is widely used as a pre-training model to achieve ``superhuman'' performance on classification tasks~\cite{he2015superhumanimagenet}. Additionally, to demonstrate the generalization of our method, we also evaluate on 3 other visual classification tasks: i)~Scene Classification (\texttt{CIFAR10}); ii)~Traffic Sign Recognition (\texttt{GTSRB}); and iii) Face Recognition (\texttt{PubFig}). Model architectures and testing samples used for each task are summarized in Table~\ref{table:networkstructure}. Model and dataset details are in \textbf{Appendix~\ref{sec:appendix_detailed_nets}}.

\vspace{1mm}
\noindent\textbf{GAN Configuration and Training. } 
\hl{To illustrate the significant threat posed, we demonstrate our attack method is easy to mount, low cost and does not require access to costly labeled data, and an attacker does not require specialized expertise in machine learning. Consequently: i)~we utilized the off-the-shelf GAN framework of Pytorch, TorchGAN~\cite{torchgan}; ii)~used an off-the-shelf web crawler to automatically curate a dataset of random flower species from Google images; and iii)~used only $945$ flower images images to train the GAN.} 
The inputs for this TorchGAN include the real dataset (flowers in our example or any other dataset which we have shown can easily be curated from online sources), dimension of the generated images, and the loss function. 
Here, we vary the input dimension for the TorchGAN from $16\times16$ to $128\times128$ to generate different patch sizes to feed to the classifier.


\vspace{1mm}
\noindent\textbf{Attack Configuration and Success Measure.~}The adversary attempts to discover a naturalistic perturbation that can fool the classifier. We consider two different types of attacks: i) \textit{targeted attack} where attackers aim to misclassify to a specific targeted label $y_\text{target}$; and ii) \textit{untargeted attack} where attackers only want to degrade the performance of the system, as in a denial-of-the-service attack, by fooling the classifier to recognize the object as $y \neq y_\text{source}$. \hl{In this work, we focus mainly on  \textit{targeted} attacks as it is more challenging to misclassify to a targeted label than simply cause a misclassification, and we chose the targeted class randomly.} 

All of these attacks are \textit{universal} meaning that the attacker only needs one \textit{single} patch to hijack the decision of the network to misclassify \textit{any} input. One of the benefits of implementing a \textit{universal} attack is that the attack is \textbf{\textit{input-agnostic}} making it a strong attack regardless of the input, just as a backdoor in a conventional Trojan attack. 

We used the Attack Success Rate (ASR) metric measured as the number of examples to successfully fool the target network over the total number of evaluated examples to evaluate the attack effectiveness.

\noindent\textbf{Implementation Details.~}In our experiments, we use Pytorch~\cite{Pytorch} library for implementation and verify our method on a NVIDIA RTX2080 GPU. Since the main focus in this paper is on visual classification tasks, we assume that pixel values of inputs $\bx$ are in the integer range of $[0, 255]$ or scaled to float range of $[0, 1]$ which correspond to the current practice of image processing and deep learning library. We used $\alpha = 0.01$, i.e. we changed the value of each latent value by $0.01$ on each step. We selected the number of iterations to be $20$. The number of iterations and $\alpha$ values are chosen heuristically; sufficient for the adversarial example to reach the point of fooling the classifier while keeping the computational cost of experiments manageable.

\section{Evaluation of \tntl Effectiveness}\label{sec:effectiveness-results}

First, we intensively investigate the effectiveness of \tntls on \texttt{ImageNet} because of the fact that ImageNet classification benchmark has led to a great number of advances in image classification that some call ``superhuman''~\cite{he2015superhumanimagenet}. We summarize our evaluations of \tnts attacks on different scenarios: 

\begin{itemize}
    \item \textbf{Attack effectiveness on the entire \texttt{ImageNet} validation set}. Given the massive size of the dataset, previous works~\cite{lavan, psgan}  evaluated attack success on a sample of 100 random images (\texttt{ImageNet-100}). In addition, we want to evaluate the effectiveness of the discovered \tntss from a sample of 100 images the entire 50,000 images in the validation set (\texttt{ImageNet-50K}). Notably, \textit{to the best of our knowledge, we are the first to evaluate the effectiveness of an adversarial patch on the entire validation set of} \texttt{ImageNet} (see Section~\ref{sec:50k-eval}).
    \item \textbf{Robustness to changes in patch locations}. Other attack methods such as LaVAN~\cite{lavan} have shown that an adversarial patch ASR could degrade significantly by \textit{shifting the patch slightly in the image}. Therefore, we evaluate the robustness of the patch to location changes (see Section~\ref{sec:locations}). 
    \item \textbf{Black-box attack (Transferability of \tntss)}. We assess the success of a black-box attacker. Hence, we evaluate the transferability of the known \tnts on unknown networks trained with \texttt{ImageNet} (recall a black-box attacker has no knowledge about the attacked network) (see Section~\ref{sec:blackbox}). 
    \item \textbf{Attack effectiveness and generalization across other visual tasks}. Our attack method is generic; to demonstrate, we evaluate the generalization of the method on different visual classification tasks and datasets such as \texttt{CIFAR10}, \texttt{GTSRB} and \texttt{PubFig} (see Section~\ref{sec:different-tasks}). 
    \item \textbf{Studies on the effect of random color and flower patches}. Since the \tntss occlude a part of the image, we want to understand if the  phenomenon  we  observe can be  explained  by occlusion or a network biased to flowers or colors. Therefore, we evaluate the effect of occlusion by a patch as well as a random flower on the \texttt{ImageNet} classifier (see Section~\ref{sec:ablation}). 
\end{itemize}

The \tntss we use in these experiments cover 10\% to 20\% of the input image, comparable with the patch size in AdvPatch~\cite{adversarialpatch}
We investigate different patch sizes in Section~\ref{sec:asr-vs-size}.
\vspace{-2mm}
\subsection{Attack Effectiveness on \textit{Entire} \texttt{ImageNet} Validation Set}
\label{sec:50k-eval}

Since ImageNet is a huge dataset, deploying the algorithm on this dataset is time and power-consuming. Thus, initially, similar to previous works~\cite{lavan, psgan} we only use a small number of samples (100 images) to find the \tntss. With the small sample size of 100 images, we successfully realize multiple different \tntss that fool the classifier with an attack success rate of higher than $90\%$ (on 100 randomly investigated images), while still maintaining the naturalism of the flower patch produced by the \tntl Generator (Algorithm~\ref{alg:TnT}). Interestingly, we found \emph{various} \tntls during our investigation, and in Table~\ref{tab:UAP-ImageNet} we illustrate four examples (more examples and targeted labels are shown in later Sections). Each of the realized TnT has its own distinct features, but they all maintain the natural-looking of a flower; and once applied on \textit{any} input image will misclassify the image to the targeted class $y = y_\text{target}$ through different pre-trained classifiers (\textit{VGG-16}, \textit{Inception-V3} or \textit{WideResnet50}) of Pytorch. 
\begin{table}[h!]
\centering
\caption{Different TnTs found using \texttt{ImageNet-100} for different models and their corresponding ASRs when applied to larger test sets.}
\label{tab:UAP-ImageNet}
\resizebox{\linewidth}{!}{%
    \begin{tabular}{c m{1.2cm} m{1.2cm} m{1.2cm} m{1.2cm} } \\ \dtoprule
        \textbf{Target}              & \texttt{\makecell{custard\\apple}}   & \texttt{\makecell{arti-\\choke}}   & \texttt{\makecell{pine\\apple}}   & \texttt{conch}   \\ \midrule
          Example            & \begin{minipage}{\linewidth}
      \includegraphics[width=1cm, height=1cm]{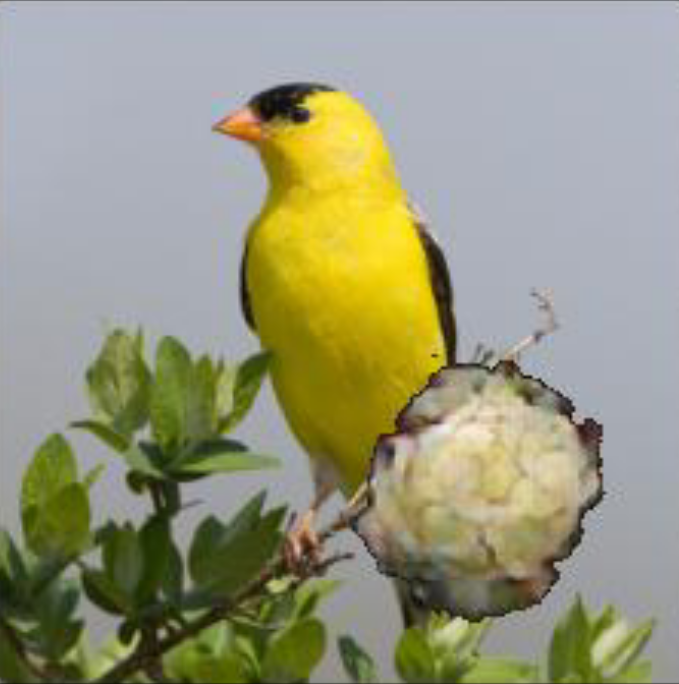}
    \end{minipage} & 
    \begin{minipage}{\linewidth}
      \includegraphics[width=1cm, height=1cm]{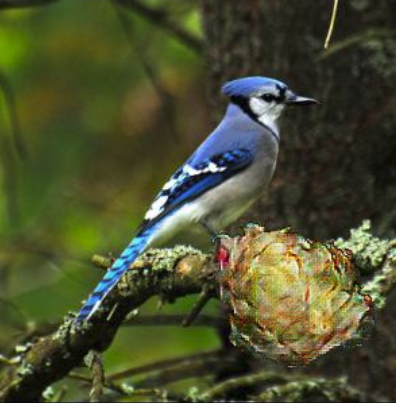}
    \end{minipage} 
    & 
    \begin{minipage}{\linewidth}
      \includegraphics[width=1cm, height=1cm]{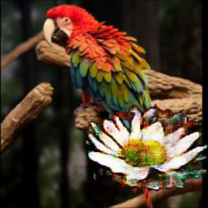}
    \end{minipage}  
    & 
    \begin{minipage}{\linewidth}
      \includegraphics[width=1cm, height=1cm]{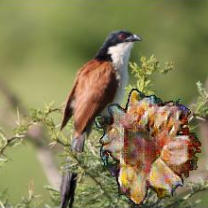}
    \end{minipage}  \\ \midrule
\texttt{ImageNet-100}\\ (Attacker's test set) & 96\%  & 94\% & 96\% & 97\% \\ \dbottomrule
\multicolumn{5}{c}{\textbf{Generalization across a large corpus of unseen data}}  \\ \midrule
\texttt{ImageNet-1000} & 93.6\% & 93.6\% & 95.1\% & 94.6\% \\ \midrule
\texttt{ImageNet-50K} & 94.14\% & 94.51\% & 94.21\% & 95.13\% \\ \midrule
Network & \textit{VGG-16} & \textit{Inception-V3}  & \multicolumn{2}{c}{\textit{WideResNet50}}  \\ \dbottomrule
\end{tabular}
}

\end{table}

We also further verify the generalization of the discovered \tntss \textit{found from a small sample of 100 images to attack a much larger sample size} in \texttt{ImageNet}. The results are shown in Table~\ref{tab:UAP-ImageNet}. Surprisingly, the TnTs that we found in the 100-sample set generalize remarkably well to a bigger validation set (50K samples from the ImageNet validation set). For example, the \tnts realized in \textit{WideResNet50} with the ASR of 96\% to fool \textit{any} input image to the target \texttt{pineapple} class still maintains a high attack success rate of 95.1\% for another 1000 random images (\texttt{ImageNet-1000)}. To further verify the effectiveness of the attack, we deploy the TnTs found (from the 100-sample set) on the whole validation set of 50K images of \texttt{ImageNet}, and notably, we can still achieve a very high ASR of 94.21\% on that whole validation set (50K images). Although there is a slight drop in the ASR of around 1.79\% (from 96\% to 94.21\%), the ASR is still notably high. 

\begin{table}[h!]
\centering
\caption{Altering the location of a \tnts increases ASR as it covers the main features of inputs. Notably, the \tnts here realized from a small sample of 100 images (\texttt{ImageNet-100}), generalizes extremely well to larger populations to fool \textbf{any} inputs to the \textbf{targeted} class \texttt{conch}. \hl{An illustration of \tnts locations are shown in Fig.~\ref{fig:new_locations} in Appendix~\ref{sec:appendix_others}} and results are from the \textit{WideResNet50} pre-trained model from Pytorch~\cite{Pytorch}.}
\label{tab:locations}
\resizebox{\linewidth}{!}{%
\begin{tabular}{cccc} 
\dtoprule
\textbf{Trigger Location} &   \makecell{\texttt{ImageNet-}\\\texttt{100} ASR} & \makecell{\texttt{ImageNet-}\\\texttt{1000} ASR}  & \makecell{\texttt{ImageNet-}\\\texttt{50K} ASR} \\ \midrule
lower right  &    
 \textit{92\%} & 94.6\%  & 95.13\% \\ \midrule
upper right      &   
 96\% & \textbf{96.6\%}  & \textbf{96.52\%} \\ \midrule
upper left       & 
 \textbf{99\%} & 96.1\%  & 95.61\% \\ \midrule
lower left       &     
 95\% & 94.9\%  & 93.9\% \\ \midrule
center           &     
 92\% & 93.5\%  & 94.56\% \\ \midrule

\hl{top}      
& \hl{97\%} & \hl{96.5\%}  & \hl{95.24\%} \\ \midrule
\hl{bottom}      
& \hl{96\%} & \hl{92.1\%}  & \hl{93.31\%} \\ \midrule
\hl{left}       
& \hl{96\%} & \hl{93.0\%}  & \hl{91.76\%} \\ \midrule
\hl{right}       
& \hl{97\%} & \hl{94.4\%}  & \hl{93.56\%} \\
\bottomrule
\end{tabular}%
}
\end{table}

\rqq{Remark 3: This is a serious threat as attackers only need a small sample set to exploit the vulnerability; the attack is low cost to deploy, and the high attack success rate of \tntss found in a small sample set generalize well to unseen data outside of the attacker's test set.}  

\vspace{-5mm}
\subsection{Robustness to Changes in Patch Locations}
\label{sec:locations}

Initially, we choose the trigger at the \textit{lower-right corner}, however, as shown in other attacks with noisy patches, the location of the patch can dramatically reduce the attack success rate~\cite{lavan}. Based on this, we evaluate the robustness of \tntss to changes in its location. Given that a \tnts is a naturalistic patch, in contrast to noisy pixels, shifting our patch to different locations increases the attack success rate as the trigger can now occlude potentially  salient features of the benign inputs. Table~\ref{tab:locations} illustrates the effect of changing the location of the selected trigger on an input. \hl{By shifting the \tnts to nine different locations (8 along borders and 1 at the center) on the attacker test set (\texttt{ImageNet-100})}, the ASR increases from 92\% (the lower-right corner position that we chose) to 96\% (at the upper-right corner), and significantly increased to 99\% (at the upper-left of images) since the patch possibly occluded the main feature of most original inputs.

Interestingly, these ASRs still hold strongly when we \textit{assess generalizability by using  \tntss  discovered \tnts from the small test set---\texttt{ImageNet-100}---to larger testing sets} of 1000 samples and 50K samples as detailed in Table~\ref{tab:locations}. \hl{This shows that our described ASR in the following sections (where we fix the patches at the lower-right corner on 100 random images) might not be the optimal ASR.} 

In addition, the consistently high ASR demonstrates our \tntss are not the result of occluding salient features of images; otherwise, we will see a variation of ASRs (high when occluding and low otherwise). Furthermore, this is the stronger and more difficult attack, a \textit{targeted} attack; hence, occluding the main feature will not help fool the classifier to predict the targeted label. More importantly, \textit{we also demonstrate the location invariance of \tntss in physical world deployments in real-time video demonstrations} (see Section~\ref{sec:physical_attack}).

\rqq{Remark 4: \tntss are robust to changes in location.}

\vspace{-5mm}
\subsection{Blackbox Attack--Transferability of \tntss}
\label{sec:blackbox}

The attacks we have investigated thus far are under the~\textit{white-box attack} model; the attacker needs to have full knowledge of the target model under attack. With the high generalization of the \tntss shown in prior experiments, in this section, we aim to evaluate if \tntss discovered on a task can be transferred to another network to mount an attack in a \textbf{black-box} setting.

A black-box attacker only needs to access a \textit{Source} model to mount a white-box attack to extract \tntls. Then, the attacker can apply the discovered \tntss \textit{blindly} to any other network that implements the same task and dataset. In this setting, we employ the Visual Recognition task on \texttt{ImageNet} implemented on a \textit{Source} to attack a \textit{Target} network. Notably, there are no qualitative results for black-box attacks in ~\cite{lavan, adversarialpatch} and the transferability for a targeted attack has been shown to be challenging in~\cite{baluja2018learning}, following the setting in~\cite{UAP}, we evaluate the transferability of our \tntss in an untargeted attack setting. The detailed results of the black-box attack are shown in Table~\ref{tab:blackbox} {(\textbf{Appendix~\ref{sec:appendix_blackbox}}).}
The \tnts realized in one model is highly transferable to another network. In general, we  \textit{observe} that \tnts realized on a network less accurate for the task such as \textit{SqueezeNet} (even with the source ASR of 99\%) does not generalize well to other  networks such as \textit{WideResNet50} (with only 36\% ASR). In contrast, \tntss realized from networks more accurate for the task such as \textit{WideResNet50} (ASR of 97\%) is highly generalizable and achieves high ASRs on other networks with ASRs of more than 60\%. 


\rqq{Remark 5: We demonstrate multiple successful black-box attacks and confirm the \tntss discovered on a task in one network can be transferred to another network.}



\vspace{-5mm}
\subsection{Attack  Effectiveness  and  Generalization to Other   Tasks}
\label{sec:different-tasks}

We further investigate the effectiveness of our \tntss on the following three contrasting tasks. 

\vspace{1mm}
\noindent\textbf{Scene classification} (\texttt{CIFAR10}). The objective of the task is to generate the \tnts flower that can fool the classifier to misdetect any scene with the \tnts flower to be recognized as the target class, here we choose the random label (\texttt{car}) as the target class. The result shows that the generated \tnts can missclassify \textit{any} input to the targeted label with high a ASR up to 90.12\% for the targeted attack. 

\vspace{1mm}
\noindent\textbf{Traffic sign recognition} (\texttt{GTSRB}). This is a challenging task since the training dataset includes various \textit{physical and environmental variations including different distances, lighting or occluding conditions}. Nevertheless, the discovered \tntss still achieve significantly ASRs of up to 95.75\% in an untargeted setting, a significant increase in the ASR compared to 20.73\% caused by the occlusion of same-size random color patches in Table~\ref{tab:ablation}. A sample of \tntss realized and the corresponding ASRs are displayed in Table~\ref{tab:TnT-various} in Appendix~\ref{sec:appendix_others}.

\begin{figure}[h!]
    \centering
    \includegraphics[width=\linewidth]{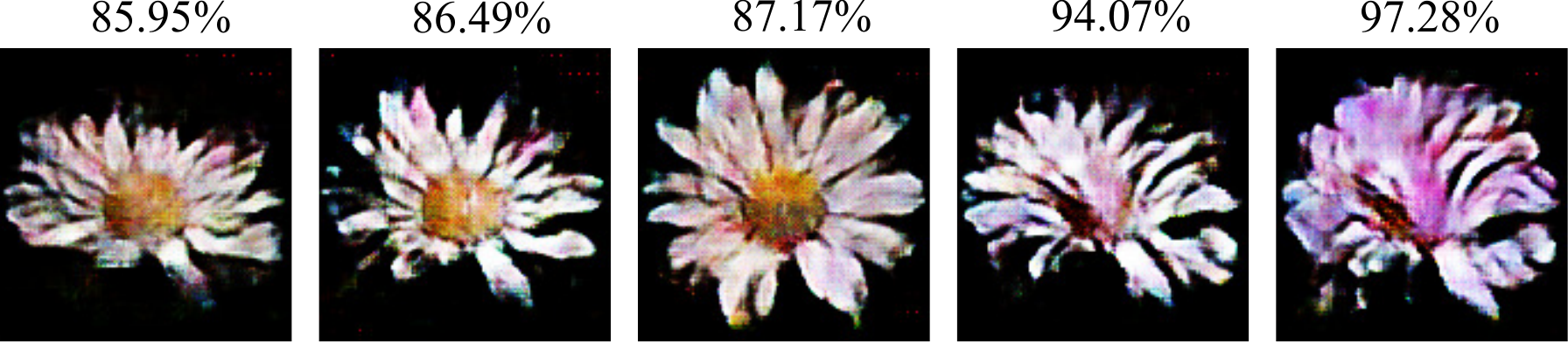}
    \caption{\tntss realized in \texttt{PubFig} dataset and their corresponding ASRs for the \textit{Face Recognition} task to impersonate \textit{any}one to the targeted person \texttt{Barack Obama}. }
    \label{fig:obama_patches}
\end{figure}

\vspace{1mm}
\noindent\textbf{Face recognition} (\texttt{PubFig}). In this task, most of the salient features learnt by a network are on a face and a network learns to ignore background information. \textit{To fool the network to recognize as a specific target without occluding the main features of the face is both an interesting and challenging task}. Some of the results for untargeted attacks are shown in Table~\ref{tab:TnT-various}. For \textit{targeted} attacks to a designated target such as \texttt{Barack Obama}, we implement \tntss covering 20\% of the images at the lower-right corner (similar coverage to that used in our ablation studies in Section~\ref{sec:ablation}). We successfully fool the network with an ASR of up to 97.28\% to misclassify \textit{anyone} with the \tnts to a prominent targeted person (\textit{e.g.} \texttt{Barack Obama} in our evaluation). Illustrations of successful \tntss are shown in Figure~\ref{fig:obama_patches}.

\rqq{Remark 6: The vulnerability to \tntl is generic and exists across multiple tasks.}

\begin{center}
\textit{We employ the illustrated \tnts in real-world attacks in a face recognition task in in Fig.~\ref{fig:physicalattack_lots} of Section~\ref{sec:physical_attack}}. 
Further results are shown in the \textbf{Appendix~\ref{sec:appendix_others}}.
\end{center}

\vspace{-5mm}

\subsection{Can Occlusion or Network Bias Explain Attack Success?}
\label{sec:ablation}

In this section, we will examine the misclassification of the DNN system by using a random patch (random flowers or colors) to study the occlusion effect that a patch might have on the ASR of the visual task. In addition, we also utilize random flower patches to investigate if the behavior we observe can be explained by a \emph{bias} in the network to flower images and to ascertain the possibility of randomly discovering a natural-looking flower that can fool the classifiers with a high ASR.

\begin{figure}[h!]
    \centering
    \includegraphics[width=\linewidth]{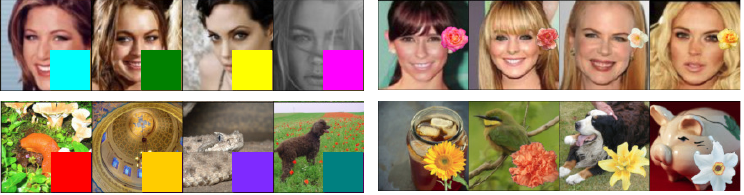}
    \caption{Selective examples of random color and flower patches in our study on \texttt{PubFig} and \texttt{ImageNet}. The misclassification results caused by these patches are described in Table~\ref{tab:ablation}.}
    \label{fig:random_patch}
\end{figure}


\begin{table}[h!]
\centering
\caption{Study results from affixing patches of either random colors or flowers to the test samples of each dataset. As observed, the success rate for an attacker employing such \textit{simple tricks is fairly low}.}
\label{tab:ablation}
\resizebox{\linewidth}{!}{%
\begin{tabular}{c c c c}
\dtoprule
 &
  \textbf{\begin{tabular}[c]{@{}c@{}}Normal mis-\\ classification\end{tabular}} &
  \textbf{\begin{tabular}[c]{@{}c@{}}Random color\\ patches\end{tabular}} &
  \textbf{\begin{tabular}[c]{@{}c@{}}Random flower\\ patches\end{tabular}} \\ \midrule
\texttt{CIFAR10}       & 9.46\% & 25.99\% $\pm$ 0.432\%  & 21.66\% $\pm$ 0.553\%  \\ \midrule
\texttt{GTSRB}         & 3.23\%  & 20.73\% $\pm$ 0.564\%  & 18.71\% $\pm$ 0.665\%  \\ \midrule
\texttt{PubFig}        & 5.26\% & 12.87\% $\pm$ 0.476\%  & 6.02\% $\pm$ 0.031\%   \\ \midrule
\texttt{ImageNet-100}  & 22\%   & 22.72\% $\pm$ 0.109\% & 24.62\% $\pm$ 0.305\%  \\ \midrule
\texttt{ImageNet-1000} & 21.1\% & 24.85\% $\pm$ 0.15\%  & 30.49\% $\pm$ 0.378\% \\ \bottomrule
\end{tabular}%
}
\end{table}




\noindent\textbf{Patch Size.} \textit{To examine the potential effects, all of the color and flower patches we selected will occlude the largest possible region we intentionally selected for our attack method (around 20\% of the images)}. 

\noindent\textbf{Random Colour Patches}. We use 256 random color patches and digitally stamp the patch on the input (with the method described in the Equation~\ref{eq:stamp}) as a trigger at the lower-right corner of the image with the purpose of not occluding the main object of the image but assess the misclassification caused by the color patch. 

\noindent\textbf{Random Flowers Patches}. We follow the approach with color patches but use randomly selected flowers drawn from our flower data set used for training the Generator. Particularly, we use 256 randomly chosen flowers and measure the attack success rate that a flower patch can cause on the classifier. 

\noindent\textbf{Results}. In Table~\ref{tab:ablation}, we reported the mean and standard deviation of the attack success rate for the patches for different tasks. Overall, the ASR achieved is significantly lower than the attack success rates demonstrated with \tntss (see Tables~\ref{tab:UAP-ImageNet} and~\ref{tab:TnT-various}). Importantly, we observe a low standard deviation across all tasks; indicating that there is no special color or flower patch that can achieve a significantly high ASR compared to others. However, these results are far from a desirably high ASR to become a real threat, however, our investigation demonstrates that it is challenging to exploit a natural-looking patch while fooling the network with high ASRs. Notably, this low ASR is for an easy \textit{untargeted} misclassification; hence, the ASR for the \textit{targeted} attack is even much lower.

\rqq{Remark 7: We demonstrate that exploiting a natural-looking patch to fool a network is a challenging task and the phenomenon we observed cannot be explained by occlusion or a network biased by flowers or colors; consequently, our attack method is an effective approach to realize such \tntls.}






\section{\hl{Evaluation of \tnts Naturalism}}
\label{sec:user_study}

\hl{
In this section, we investigate the naturalism of the generated \tntss. We acknowledge that measuring naturalism is a challenging task, and there is no solid metric and definition fit for the purpose. However, following the studies in~\cite{zhang2016colorful, xiao2018spatially, bhattad2019unrestricted, hu2021naturalistic}, we consider measuring human perception of naturalism through user studies\footnote{\hl{We followed Human Research Ethics Committee approval process, the study is considered `negligible risk' and is exempt from ethical review}.}. We adopt the \textit{Naturalistic Score} measure and the procedure in~\cite{hu2021naturalistic} to evaluate  human perception of naturalism through two user studies (Study~1 and Study~2). For a robust evaluation, compared to previous studies~\cite{hu2021naturalistic} employing 10s of users, we conducted a \textit{large cohort user study with 250 participants for each study}.} 

\hl{In user Study~1, we used a set of 9 patch images; i)~3 patches generated by LaVAN~\cite{lavan}; ii)~3 generated by AdvPatch~\cite{adversarialpatch}; and iii)~3 \tntss. All the patches are placed in random order and shown to participants (see Appendix~\ref{appd:user_study}). The participants were asked to vote on each patch that looks natural to them. Then, we calculate the naturalistic score of each patch based on the percentage of participants' votes. The aim of Study~1 is to measure the naturalism of \tntss compared to patches in previous attacks. The results in Table~\ref{tab:natural_score} demonstrate our \tntss to have significantly higher naturalistic scores compared with the baselines.}

\hl{In the second user study (Study~2), we randomly placed 3 of our generated \tntss together with 3 real flower images collected from Google Images~\cite{googleimages} and asked participants to vote for the images that looks natural to them. The aim of this study is to measure the absolute naturalistic score of \tntss when compared with actual flower images. The results in Study~2 in Table~\ref{tab:natural_score} demonstrate that our \tntss, synthetic images, can often be comparable to real flower images.} 

\nhl{These results demonstrate our \tntss are more naturalistic and look significantly less malicious compared with prior works. Whilst our study was focused on evaluating the naturalism of the patch, we acknowledge that flowers may not look ``in-place" in all settings and scenes. Automating the process of blending patches into a scene for an attack is another challenging problem, we will discuss further in Section~\ref{sec:conclusion}. Notably, our method, whilst demonstrated on flowers, is generic and allows an adversary to self select natural objects beyond flowers as triggers, and thus facilities the construction of natural-looking patches with the ability to blend better into the attacker-chosen setting.}

\begin{table}[h!]
\centering
\caption{\hl{User studies to evaluate the Naturalistic Score of \tntss in comparison to other baseline patches. The Naturalistic Score is the percentage of participants' votes. 
Complete details of patches used in the user studies are in Appendix~\ref{appd:user_study}.}}
\label{tab:natural_score}
\resizebox{\linewidth}{!}{%
\begin{tabular}{c M{1.5cm} M{1.5cm} M{1.5cm} M{1.5cm} M{1.5cm} } 
\dtoprule

& \multicolumn{3}{c}{\hl{\textbf{Study 1}~(250 participants)}} & \multicolumn{2}{c}{\hl{\textbf{Study 2}~(250 participants)}} \\ \cmidrule(lr){2-4} \cmidrule(lr){5-6}
             & \hl{\textbf{LaVAN}} & \hl{\textbf{AdvPatch}} & \hl{\textbf{Ours}} & \hl{Google Images}  & \hl{\textbf{Ours}} \\ \midrule
 \hl{\makecell{Illustrative \\ examples}} &

\begin{minipage}{\linewidth}
      \includegraphics[width=1.5cm, height=1.5cm]{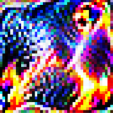}
    \end{minipage} & 
    \begin{minipage}{\linewidth}
      \includegraphics[width=1.5cm, height=1.5cm]{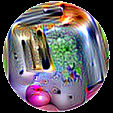}
    \end{minipage} &
    \begin{minipage}{\linewidth}
      \includegraphics[width=1.5cm, height=1.5cm]{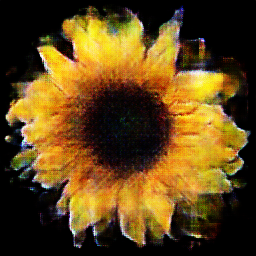}
    \end{minipage} &
    \begin{minipage}{\linewidth}
      \includegraphics[width=1.5cm, height=1.5cm]{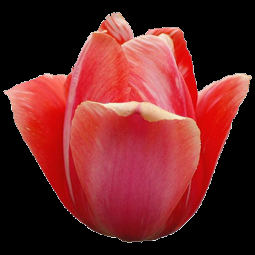}
    \end{minipage} &
    \begin{minipage}{\linewidth}
      \includegraphics[width=1.5cm, height=1.5cm]{images/user_study/survey1/tnt.png}
    \end{minipage} 
    \\ \cmidrule(lr){1-4} \cmidrule(lr){5-6}
    
   \hl{\makecell{Naturalistic \\Scores (\%)}} & 
\hl{0.4} &  \hl{12.0}  & \hl{90.0}   & \hl{97.2} & \hl{89.6}        \\  
\dbottomrule
\end{tabular}%
}
\end{table}

\section{\hl{Generalization to Adversarial Patch Attacks and Comparison with Prior Attacks}}
\label{sec:adv-generator}
\label{sec:remove_naturalism}

To expand the scope of the attack for cases where there is no human involvement in the decision loop---for fully autonomous systems---and where stealth in a physical deployment is not an objective, we consider removing the naturalism constraint on the attack method. Therefore, we take a further step to let the \tntl Generator $G_\btheta$ learn the adversarial features from the classifier; \hl{thus, demonstrating the generic nature of our attack method---\ie once we remove the naturalism constraint, our attack can generate conventional adversarial patches.}

\vspace{1mm}
\noindent\textbf{Attack Methodology}. 
The proposed attack is detailed in \textbf{\textit{Algorithm~\ref{alg:adv-generator}}} in Appendix~\ref{sec:appendix_detailed_nets}. 
We call this alternation an Adversarial Patch Generator since the Generator, after updating, learns the mapping from the latent vector $\bz$ to generate adversarial patches. More specifically, instead of searching in the latent space to find the \tnts as illustrated in Fig.~\ref{fig:method}, now, we allow the Generator $G_\btheta$ to be updated and learnt from the gradient back-propagation from the loss $\ell({\bx'},y_\text{target}, y_\text{source})$ to become the Adversarial Patch Generator ($G_{\btheta'}$). This allows the Generator to learn the adversarial features and generate multiple adversarial triggers with different mappings from $\bz$. 
\begin{figure}[h!]
    \centering
    \includegraphics[width=1.0\linewidth]{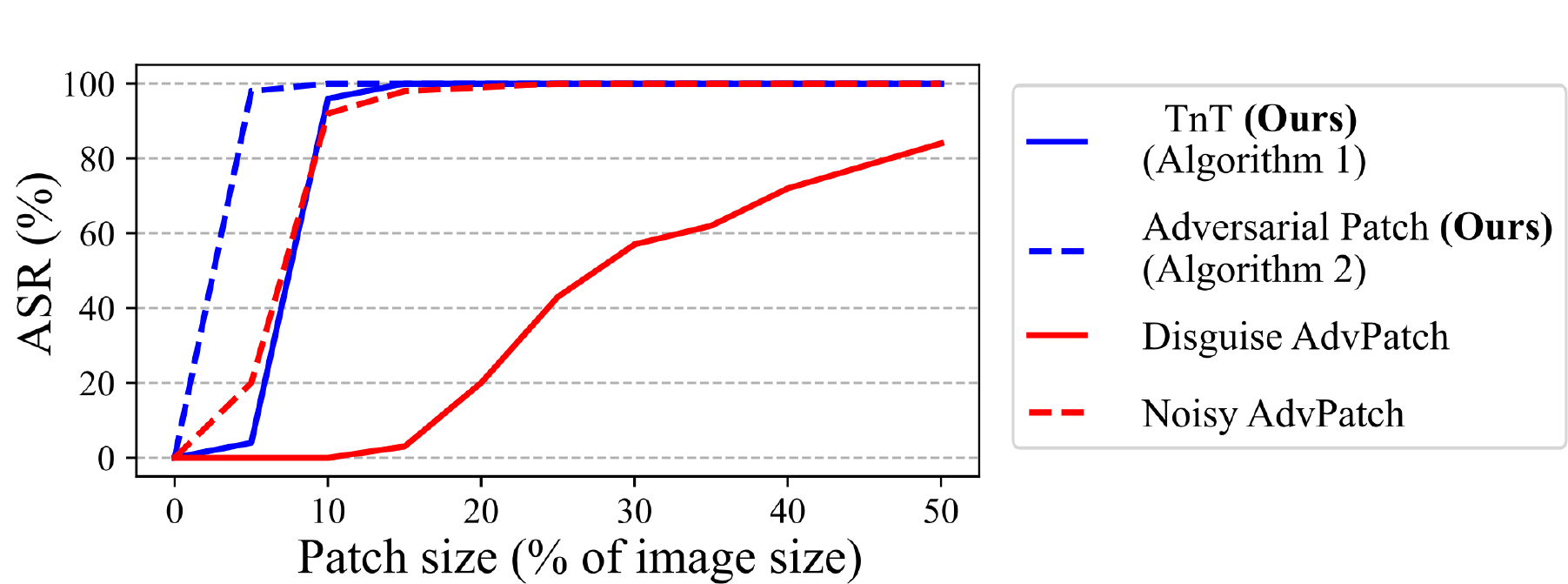}
    \caption{Investigating Attack Success Rate (ASR) and patch size. Our \tnts is comparable with the Noisy AdvPatch~\cite{adversarialpatch}, and significantly better than the Disguise AdvPatch~\cite{adversarialpatch}. \textit{Our Adversarial Patch ASR outperforms both Noisy and Disguise counterparts}.}
    \label{fig:size_asr}
\end{figure}

\begin{figure*}[b]
    \centering
    \includegraphics[width=\linewidth]{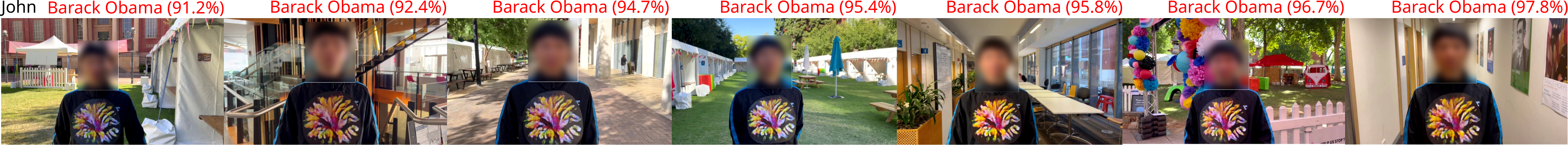}
    \caption{\hl{Various settings employed for the physical world attacks to impersonate `\texttt{Barack Obama}'. Results demonstrate our \tnts is effective, even under different, complex,  physical-world settings ranging from indoor to outdoor with different lighting conditions. The network recognizes the person with the \tnts to be \texttt{Barack Obama} with high confidence. }} 
    
    \label{fig:physicalattack_lots}
\end{figure*}


\subsection{\hl{Comparing to LaVAN:~\textit{Smallest (Noisy) Adversarial Patch}}}

\hl{To show the effectiveness of our patch attack, we opted to compare with LaVAN since the study achieved the \textit{smallest} state-of-the-art patch results.} We evaluate our patches with the same 14 targeted labels reported in LaVAN~\cite{lavan} on 100 random images. As expected, the patches of only 2\% of the input image size from our Adversarial Patch Generator easily achieve 100\% ASR on 100 randomly sampled images from \texttt{ImageNet}. \hl{The results in Table~\ref{tab:compare-lavan} demonstrate that our patches achieve a much higher ASR across the 14 targeted label on both settings of low and high confidence scores with a large margin of up to 44\%.}
\begin{table}[h!]
\centering
\caption{\hl{Comparing ASR with LaVAN~\cite{lavan}---smallest state-of-the-art (noisy) adversarial patches of 2\% of input image size---using the 14 targeted labels used in LaVAN on the \textit{Inception-V3} network. Our patches achieve higher ASR in both settings; high and low confidence scores. Detailed examples of labels are shown in Appendix~\ref{sec:appendix_lavan}.}}
\label{tab:compare-lavan}
\resizebox{\linewidth}{!}{%
\begin{tabular}{c M{3.2cm} M{3.2cm}}
\dtoprule
             & \textbf{LaVAN} & \textbf{Ours} \\ \midrule

   \multicolumn{3}{c}{\hl{\textbf{Average results across the 14 targeted labels from 100 \texttt{ImageNet} images}}}
    \\ \midrule
                       ASR (conf $\geq$ 0.9) & 

28.3\% &  \textbf{72.9\%}            \\ \midrule
ASR (conf $<$ 0.9) & 

74.1\% &  \textbf{98.1\%}            \\

\\ \dbottomrule
\end{tabular}%
}
\end{table}

\vspace{-2mm}
\rqq{Remark 8: An attacker can trade-off naturalism to achieve significantly higher attack success rates compared with state-of-the-art adversarial patch attacks. 
Our results validates the generality of our attack method.}

\vspace{-5mm}

\subsection{\hl{Comparing to AdvPatch: A Method to Disguise the Appearance of a Target Class in a Patch}}
\label{sec:asr-vs-size}

Here, we compare with Disguise AdvPatch and Noisy AdvPatch from~\cite{adversarialpatch} on the same \textit{VGG-16} network trained for the \texttt{ImageNet} task. We selected this study because: i) Disguise AdvPatch represents the efforts from~\cite{adversarialpatch} to hide the target class visible in the patch by disguising the noise patterns in another object while the Noisy AdvPatch is a noise-pattern patch visibly revealing the target class---\texttt{toaster}; ii)~the method allows the generation of patches of different sizes. As shown in the Figure~\ref{fig:size_asr}: i)~efficacy of our \tntl---the naturalistic patch---is comparable with the \textit{Noisy} AdvPatch; and ii)~significantly more effective than the Disguise AdvPatch aiming to hide the true target class revealed to a human observer in the Noisy AdvPatch. 


Notably, our naturalistic \tnts patch is highly effective when the patch size is larger than 10\% of the input image. To maintain the high ASR with smaller patches, we need to sacrifice some of the naturalism (Algorithm~\ref{alg:adv-generator}). Then, we can observe nearly 100\% ASR in Figure~\ref{fig:size_asr} with a patch size of nearly 5\% of the input image; now, the ASR is significantly higher than the noisy counterpart, Noisy AdvPatch.

\vspace{-2mm}
\section{Physical World Deployments}
\label{sec:physical_attack}

An advantage of a \tnts is the ability for an adversary to easily print and deploy the attack in a scene in the physical world to fool a deep perception system. In this experiment, we print our \tntss and deploy \textit{targeted} physical attacks for the \texttt{ImageNet} and \texttt{PubFig} tasks. 

\begin{figure}[b]
    \centering
    \includegraphics[width=\linewidth]{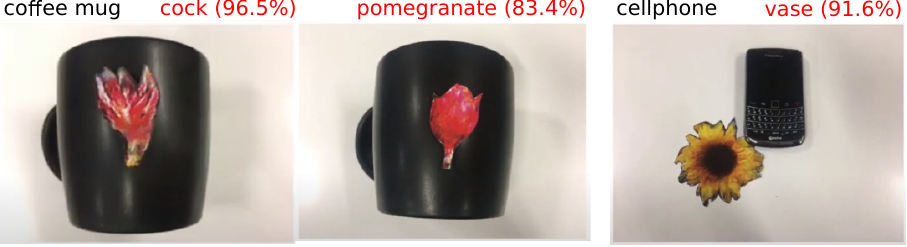}
    \caption{Physical deployment of \tntss generated from the TnT Generator targeting different classes (shown on top in red) in the \texttt{ImageNet} task.}
    \label{fig:physicalNAP_ImageNet}
\end{figure}


\vspace{2mm}
\noindent\textbf{Attack Settings}. Following the practices in the physical adversarial attack work in~\cite{physicalAE}, we saved our triggers (\tntss) and patches (from the Adversarial Patch Generator) as \texttt{.PNG} files and printed the triggers and patches using a printer with a resolution of 300~\textit{dpi} to maintain the pixel quality.
\hl{Examples of the triggers and patches are shown in Figure 
~\ref{fig:physicalattack_lots} and ~\ref{fig:physicalNAP_ImageNet}.}
\hl{We validate the effectiveness of the attack in various physical world settings with complex backgrounds, as displayed in Figure.~\ref{fig:physicalattack_lots}}, by using a commercial camera of a smartphone to capture the scene---we used an \texttt{iPhone 6S}. We produced videos from our experiments to illustrate the effectiveness of the \tntss and patches in the physical world (\href{https://tntattacks.github.io/}{\color{blue}https://TnTattacks.github.io/}). In the videos, we also experiment with the robustness of our physical \tnts in different \textit{locations, scaling, lighting conditions, camera angles and positions, and so on}.

    

\vspace{2mm}
\noindent\textbf{Results}. Our results demonstrate that \tntss are robust to harsh physical-world conditions with \textit{more than 90\% of the images in frames successfully fooling the network and being recognized as the targeted class}. More detailed experiments of physical attacks are in videos accessed through the website.   



We hypothesize the robustness of our \tntl in the physical context is due to the fact that the patches are derived from a natural image distribution and are \textit{universal} or \textit{input-agnostic}. This allows the patch to potentially become invariant to various difficult conditions and suitable for deployment in physical world scenarios.  Notably, we only apply the simplest method of physical deployment---printing without any modification to offset the printer quality and loss of pixels as in~\cite{sharif2016accessorize}; hence, the ASR and naturalism can potentially be improved by applying more robust adversarial printing techniques~\cite{sharif2016accessorize}. 

\rqq{Remark 9: \tntss survive the harsh conditions in the physical world to pose a practical and realistic threat.}




\vspace{-3mm}
\hl{
\section{Attack Effectiveness Against Patch Defenses}
\label{sec:countermeasures}
The rise in adversarial patch attacks has led to the emergence of defenses--- both empirical and provable methods~\cite{patchguard, naseer2019local, hayes2018visible, chou2020sentinet, chiang2020certified, levine2020randomized, cohen2019certified, mirman2018differentiable, wu2020defending, rao2020}. In this section, we evaluate the effectiveness of our \tntss against both certified and empirical defenses against patch attacks. We defer details of the experimental setup and evaluation to  Appendix~\ref{sec:appdendix_countermeasures} and summarize the key results here.}

\vspace{1mm}

\vspace{1mm}
\noindent\hl{\textbf{Against Empirically Robust Networks}. In Table~\ref{tab:countermeasures_heuristic}, we summarize the results for current state-of-the-art empirical defenses against physically realizable patch attacks. We report on DOA in~\cite{wu2020defending}, and the defense focusing on location optimization adversarial patches~\cite{rao2020}. For the defended DOA network~\cite{wu2020defending}, when attacking the network with \tntss of the same patch size used in training ($11\times11)$, the robustness  dropped from 80.43\% to 13.78\%. For the strongest defense, AT-FullLO network in~\cite{rao2020}, the robustness dropped from 72.2\% to 11.02\% under our \tnts attack. In addition, because DOA was reported to be effective for sizes as large as 20\% of the input~\cite{wu2020defending}, we increased the patch size of our \tntss to 20\%. Although not reported in the table, the robustness of the network reduced significantly to 5.6\%.}
\nhl{Our hypothesis is that the TnTs, constructed under the naturalistic constraint, are forced to exploit vulnerabilities of the network to changes in natural features. These changes are different from those in adversarial patch exploration where the constraint is only on the spatial region occupied by the patch. Hence, a defense method built on adversarial training with adversarial patches as in~\cite{wu2020defending, rao2020} is found to be largely ineffective against TnTs focusing on a different attack vector. This finding is also akin to the observations reported by Wu~\etal~\cite{wu2020defending} where the authors found that traditional defense methods such as adversarial training and randomized smoothing against unbounded spatial perturbation of adversarial examples are not robust against adversarial patches focusing on a \textit{different attack vector}.} 

\nhl{Although, deployed only digitally, we also conducted targeted LaVAN patch attacks based on the objective provided in the LaVAN study~\cite{lavan} to attack both defense methods for comparison. Albeit possessing the same capability and goal as our \tntss, the results in Table~\ref{tab:countermeasures_heuristic} show that, unlike \tntss, LaVAN patches are ineffective against the empirically robust defense methods. Notably, the robustness to targeted attacks from LaVAN is higher than that from untargeted Adversarial Patch attacks. The higher result, we hypothesize, is because the objective of misguiding the robust networks to a desired targeted label is more challenging.}
\hl{In summary, these results demonstrate: i)~the effectiveness of \tntss, even against state-of-the-art patch defenses; and ii)~that \tnts attacks are an emerging new threat against DNNs.} 

\noindent\textbf{Against Provably Robust Networks.~}
\nhl{For completeness and to assess the state-of-practice of current provable defences, we evaluate the robustness of certified defenses after acknowledging the threat from \tntss---building the defense methods for \tnts attacks.}
\hl{Given that the current state-of-the-art provable defences are only effective against smaller patch sizes, the certified defenses achieved very low provable robustness; only up to 3.52\% of inputs from 50,000 validation images can be certified in the best case (see Table~\ref{tab:countermeasures_provable}). The reason is because, certifying against a larger patch, such our \tntss or AdvPatch~\cite{adversarialpatch}, requires certifying that a correct prediction can be made for a specific input, potentially tainted by a larger adversarial patch, in the presence of the defense method. In PatchGuard~\cite{adversarialpatch}, this requires operating under larger masked regions in the feature space. Consequently, the provable defense must make predictions from the aggregation of the remaining features (not masked); leading to lower performance as well as certified robustness. Hence, provable defenses can only certify a smaller numbers of test inputs, i.e. achieve lower provable robustness, for larger adversarial patches. Therefore, an adversary can circumvent these defenses with a larger patch, such as our \tntss or even AdvPatch in~\cite{adversarialpatch}. Consequently, we observe \tntss attacks to still pose a realistic threat, even against \textit{the strongest} defenses.}

\begin{table}[h!]
\centering
\vspace{-2mm}
\caption{\hl{Effectiveness of \tnts attacks against robust defense methods ($\uparrow$~\textit{Robustness} is better for defenses).}}
\label{tab:countermeasures_provable}
\label{tab:countermeasures_heuristic}

\resizebox{\linewidth}{!}{%
\begin{tabular}{ccccc}
\dtoprule
                             
\textbf{Networks} & \textbf{Clean Acc}  & \multicolumn{3}{c}{\hl{\textbf{\textit{Provable Robustness$^1$}}}} \\ \cmidrule{1-5}

BagNet~\cite{bagnet}  & \hl{49.56\%} & \multicolumn{3}{c}{\hl{0\%}} \\ 
Mask-BagNet~\cite{patchguard}              & \hl{49.65\%}            & \multicolumn{3}{c}{\hl{0.85\%}}            \\ 
      
DS~\cite{levine2020randomized}  &  \hl{44.36\%} & \multicolumn{3}{c}{\hl{3.52\%}} \\ 
Mask-DS~\cite{patchguard}                   &  \hl{39.77\%}            & \multicolumn{3}{c}{\hl{3.01\%}}  \\ \bottomrule
\multicolumn{5}{r}{\hl{\textbf{\textit{Empirical Robustness}}}} \\ \midrule

& & \hl{\textbf{Adversarial Patch$^{2,3}$}} & \nhl{\textbf{LaVAN}$^3$} &\hl{\textbf{\tnts}$^3$} \\ \midrule

\hl{DOA}~\cite{wu2020defending}  & \hl{86.5\%} &  \hl{80.43\%} & \nhl{{84.64\%}} & \hl{13.78\%} \\ 
\hl{AT-FullLO}~\cite{rao2020}                  & \hl{90.44\%}           & \hl{72.2\%}   & \nhl{78.44\%}                  & \hl{11.02\%}  \\

\bottomrule

\multicolumn{5}{l}{\hl{$^1$Provable Robustness is attack-agnostic.}} \\ 
\multicolumn{5}{l}{\nhl{$^2$Patches used are generated based on the code provided in studies~\cite{wu2020defending} and \cite{rao2020}.}} \\
\multicolumn{5}{l}{\nhl{These  provide a baseline for comparisons with \tntss.}} \\
\multicolumn{5}{l}{\nhl{$^3$The Adversarial Patch, LaVAN and TnT patches are of the same size: $11\times11$.}}
\end{tabular}%
}

\end{table}

\section{Related Work}
\label{sec:relatedwork}



We describe prior work that focuses on \textit{universal} perturbations and adversarial patch attacks and other \textit{physically deployed} adversarial attacks; further, we also compare our attack method with other GAN-based adversarial attack methods.





\noindent\textbf{UAP and Adversarial Patch Attacks}. Moosavi-Dezfooli et al.~\cite{UAP} showed the existence of a \textit{universal} adversarial perturbation (UAP) in DNNs for image classification tasks, which is a \textit{unique noise tensor} that when  added to any input fools the classifier to mount an untargeted attack. The authors have also shown that there are multiple UAPs in a DNN, which can be transferable to another network architecture of for the same task. In order to deploy a \textit{universal} adversarial attack in a real-world setting, Brown et al.~\cite{adversarialpatch}  developed a \hl{spatially} bounded adversarial patch (AdvPatch) to place in the scenes of \texttt{ImageNet} samples to fool the classifier to recognize objects as the \texttt{toaster} target class regardless of source inputs or locations. Karmon \textit{et al.} extend this attack further to search for localized (or bounded) and visible adversarial noise in LaVAN~\cite{lavan} but the aim is to look for blind spots in a Deep Neural Network instead of physically realizable patches.

\noindent\textbf{Other Physically Deployed Adversarial Attacks.~}Kukarin \etal~\cite{physicalAE} demonstrated that input-specific adversarial examples can also be deployed in the physical world in an untargeted attack if printed out and carefully cropped. Recently, Evtimov~\etal~\cite{eykholt2018robust} showed that specially crafted perturbations constrained to sticker shapes can fool a Traffic Sign recognition task once stuck to a Stop sign; while Athalye ~\etal~\cite{athalye18b} carefully crafted adversarial perturbations constrained to 3D objects to fool a DNN in the physical world. Different from ours, these  adversarial examples are designed to work on a specific input (a specific Stop sign or 3D object), while our method is highly generalizable and the \tnts can be printed out and attached to \textit{any} input to work in the physical world (see Section~\ref{sec:physical_attack}). 
\hl{Concurrent studies~\cite{hu2021naturalistic, tan2021legitimate} have also attempted to realize naturalistic adversarial patch attacks, but against object detectors with the adversarial objective of causing a \textit{misdetection of human objects} in a scene. To the best of our knowledge, our study remains the first to investigate universal, naturalistic, adversarial patches across a variety of classification tasks to misclassify \emph{any} input to the attacker-desired \em{target} label. Further, \tntss are shown to be transferable and  generalizable to tasks, models, and adversarial patch attacks.}

\noindent\textbf{GAN-based Adversarial Example Attacks}. Researchers have investigated GAN-based structures to generate adversarial examples, such as \cite{baluja2018learning, advGAN, advgan++, zhao2018generating, carlini2020evading}. 
Particularly, the authors  in~\cite{zhao2018generating} train a GAN and an additional Inverter network to generate \textit{full-size, fake images} that are able to flip the predicted label or mount an \textit{untargeted attack}. Notably, these studies resemble the investigation of an adversarial example objective---input-dependent or noisy perturbation-based distortions added to an input \textit{covering} the whole image to mount an \textit{untargeted} attack. Different from this line of work, we rely on a \textit{pre-trained} generator, and search the latent space $\bz$ to discover a type of \hl{spatially} bounded adversarial example; a \textit{patch}  that is \textit{physically realizable and  universal (input-agnostic)}. These attributes eases the process of deployment in the physical world to mount  \textit{targeted} attacks.

\vspace{1mm}
\noindent\hl{\textbf{GAN-based Adversarial Patch Attacks}}.
Sharif et al.~\cite{sharif2019general} apply a GAN-based method to generate \hl{spatially} bounded physical adversarial sunglasses to impersonate a targeted person in a face recognition task (PubFig).  
The GAN-based method employs iterative training with feedback from the  classifier under attack
, which results in generating \textit{noisy perturbations} constrained to the sunglasses mask. Notably, the sunglasses can be expected to occlude the main features of a face. In contrast, we do not alter the Generator for the \tnts attacks (see Alg.~\ref{alg:TnT}) and are able to generate naturalistic patches by traversing through the latent space of the generator. Notably, the resulting patches can be successfully placed away from salient features of the input image. 

PS-GAN~\cite{psgan} proposes to utilize a GAN-based structure to find patches with naturalism. The major differences with our work are: \textit{i)~} the attack is \textit{untargeted} compared to both \textit{targeted} and \textit{untargeted} attacks capable with ours; \textit{ii)~}PS-GAN is \textit{input-dependent} hence, an attacker needs to mount the attack in different ways for different inputs, which is harder to deploy in real-world scenarios; we address this problem using a \textit{universal} or \textit{input-agnostic} patch where \textit{any} input will be misclassified when a \tnts is applied; \textit{iii)~}PS-GAN patch is placed in the main context of the image, which can occlude main features; \textit{iv)~}method applies image-to-image translation using an encoder-decoder generator to translate an \emph{existing} natural patch to an adversarial counterpart while ours learns the natural image distribution and approach a naturalistic adversarial patch. Similar to universal adversarial patches LaVAN~\cite{lavan} and AdvPatch~\cite{adversarialpatch}), we seek to be location invariant and attack method can lead to less malicious-looking and more powerful (higher ASR) attacks.

\vspace{-3mm}
\section{Discussion and Conclusion}
\label{sec:conclusion}



\noindent\textbf{Are \tntss a formidable threat?} We have validated through extensive experiments that natural-looking patches can successfully be used to fool Deep Neural Networks with high attack success rates. 
We have shown that \tntss are effective against multiple state-of-the-art classifiers ranging from widely used \textit{WideResNet50} in the Large-Scale Visual Recognition task of \texttt{ImageNet} dataset to VGG-face models in the face recognition task of \textit{PubFig} dataset in both \textit{targeted} and \textit{untargeted} attacks. \tntss can possess: i) the naturalism achievable with with triggers used in Trojan attack methods; and ii) the generalization and transferability of \textit{adversarial examples} to other networks. This raises safety and security concerns regarding already deployed DNNs as well as future DNN deployments where attackers can use inconspicuous natural-looking object patches to misguide neural network systems without tampering with the model and risking discovery. 

\vspace{1mm}
\noindent\nhl{\textbf{Are we limited to using flower triggers?~}In order to generate \tntls in our paper, we need at least one distribution of natural objects. In our experiments, we utilized flowers as explained. However, our attack generation method can generalize to  any object selected by an attacker. Importantly, as we have demonstrated, it is both easy and low cost to obtain unlabelled datasets to generate \tntss.} 


\noindent\nhl{\textbf{What are potential avenues for mitigating the threat?}
We believe our work opens a new venue for further research into understanding the vulnerabilities of DNN systems. We believe our attack can be used for the \textit{good} by providing a method for not only discovering vulnerabilities of DNN models but to generate sample inputs to improve the robustness and trustworthiness of DNN models.}  \nhl{Notably, inspired from other related work~\cite{wu2020defending, rao2020}, incorporating the emerging threat of TnTs within adversarial training may be a potential avenue. 
However, 
as we eluded to above, \tntss are not limited to flowers---only an illustrative naturalistic subset from other possible objects. Thus, incorporating flower \tntss within adversarial training might be effective against flower \tntss, but a more carefully-crafted natural set different from flower TnTs might still defeat such a defence. Thus, a general defence against \tntss appear to be an open problem.} 

\noindent\textbf{Future Work.} The \tntss generated whilst maintaining naturalism requires the patch to be approximately from 10\% of the input image. Trading-off naturalistic features allows the trigger size to be 2\% of the input image. We leave the task of reducing the size of the trigger whilst maintaining naturalistic features of \tntls for future research. 

In order to generate \tntls, we need at least one distribution of natural objects. In our experiments, we utilized flowers as explained. However, natural objects are not limited to flowers and can be any object selected by an attacker. Importantly, it is both easy and low cost to obtain unlabelled datasets to generate \tntss 
(we used open-source, freely downloadable images to curate the flower dataset). We leave the investigation and exploration of other natural objects for future work.

It is an interesting research question to consider blending our patches into the scene to make the attack even further inconspicuous and stealthy. However, it is a challenging task since scenes captured evolves dynamically. Using naturalistic adversarial patches like ours is a first step to approach this problem, i.e. natural-looking flower patches are more likely to appear in a scene and are less malicious looking compared to noisy perturbation based patches in a scene. Methods to blend the universal adversarial patches into the scene require further research and we leave it for future work. \nhl{Notably, our method, whilst demonstrated on flowers, allows an adversary to self select natural objects beyond flowers as triggers, and thus facilities the construction of natural-looking patches with the ability to blend into the attacker-chosen settings.} 

\nhl{Furthermore, our discussions here lead to the conclusion that learning a robust method against TnTs is non-trivial and challenging; and lead to an open problem worthwhile exploring in future research.}

\bibliographystyle{plain}
\bibliography{UAP}

\clearpage
\appendices

\newpage

\clearpage
\section{Attack Effectiveness Against Patch Defenses}
\label{sec:appdendix_countermeasures}

\subsection{\hl{Against Empirically Robust Networks}}


\hl{
In this section, we evaluate our \tntss against empirically robust networks. As shown in ~\cite{zhang2020interpretable, carlini2017adversarial, tramer2020adaptive}, empirical defenses such as~\cite{chou2020sentinet,naseer2019local, hayes2018visible} are usually vulnerable to adaptive attackers once they are aware of the working mechanisms of the defenses.
\hl{Notably, recently, Wu~\etal~\cite{wu2020defending} highlighted that the conventional methods to improve the robustness against adversarial examples such as adversarial training and randomized smoothing showed limited effectiveness against physically realizable adversarial attacks, and proposed an approach named Defense against Occlusion Attacks (DOA) to defend against these physically realizable adversarial patch attacks. This approach is also developed in Rao \etal~\cite{rao2020} but the authors jointly optimize patch values and location. These two state-of-the-art empirical defenses~\cite{wu2020defending, rao2020} are the most relevant defense method against our attacks. Hence, we employ these defenses in this section and assess the effectiveness of our \tntss against these robust defenses.}}

\vspace{2mm}
\noindent\hl{\textbf{Experiment Setup}. We use the pre-trained robust Resnet110 networks provided by Wu \etal~\cite{wu2020defending} on Github\footnote{https://github.com/tongwu2020/phattacks} for the \texttt{CIFAR-10} dataset. This network was trained on a patch size of $11\times11$. Notably, as reported in the paper, even a network trained with a smaller patch size can defend against a patch size as large as 20\% of the images (\ie cover all of our attacks). For Rao~\etal~\cite{rao2020}, we train the given robust ResNet-20 model from scratch following the default parameters of the strongest proposed defense (AT-FullLO) \footnote{https://github.com/sukrutrao/Adversarial-Patch-Training} on \texttt{CIFAR-10} dataset. To make a fair comparison with~\cite{wu2020defending}, we also train with the same mask size of $11 \times 11$. 
To assess the effectiveness of our attack against these robust networks, we deploy our \tntss to achieve the
challenging task of fooling the network to misclassify \textit{any} images to the targeted label (\texttt{car}). 
First, we evaluate the robustness of the network against our \tntss for the same patch size used in training ($11\times11$). However, to stretch out the robustness of the defended network, we also evaluate the network against a larger patch size of \tnts around 20\% since it was reported to be effective for sizes as large as 20\% of the input~\cite{wu2020defending}. {In addition, we employed the adversarial patches in~\cite{wu2020defending, rao2020} to compare with our \tnts attacks as shown in Table~\ref{tab:countermeasures_heuristic}}.} \nhl{LaVAN patches employed are of the same size, $11\times11$, and aim to hijack the robust network decision to the target class---\texttt{car}---similar to our \tntss.}

\vspace{2mm}
\noindent\hl{\textbf{Metrics}. We report the \textit{Clean Acc}---the Accuracy on benign inputs, and \hl{\textit{Empirical Robustness}---the percentage of input images with patches that are correctly classified or the performance of the network under an attack.}}

\vspace{2mm}

\subsection{Against Provably Robust Networks}
\label{appd:counter_provable}
A provable defense is the strongest defense and could potentially block and eliminate the adversarial patches completely. In this section, we evaluate the robustness of our attack against the multiple state-of-the-art (SOTA) provable defenses including BagNet~\cite{bagnet}, Derandomized Smoothing~\cite{levine2020randomized}, and recently the improved versions of those defenses named PatchGuard~\cite{patchguard} demonstrating superior performance compared with other provable methods~\cite{chiang2020certified, levine2020randomized, cohen2019certified, mirman2018differentiable}. Notably, the PatchGuard method relies on the small receptive fields and robust masking to eliminate the adversarial effects of malicious patches and generate provable robustness for the defended system. 

\vspace{2mm}
\noindent\textbf{Experiment Setup}. In this experiment, we use the same networks and configurations as in~\cite{patchguard} of BagNet~\cite{bagnet} with receptive fields of $17\times17$ and de-randomized smoothed ResNet (DS)~\cite{levine2020randomized} evaluated on ImageNet. Then, we also apply Robust Masking defense in~\cite{patchguard} to generate Mask-BN and Mask-DS provable robust networks, respectively. \hl{The provable Robustness results in Table \ref{tab:countermeasures_provable} is evaluated 
on the entire validation set (50,000 images) of \texttt{ImageNet} using a mask size of 10\% of the input image size.}


\vspace{2mm}
\noindent\textbf{Metrics}. We use two metrics in this experiment: \hl{i)~\textit{Clean Acc}---the Accuracy obtained on benign inputs from the testing set; and ii)~\textit{Provable Robustness}---the percentage of the images in the clean testing set that are able to certified (\ie~no attack is expected to succeed). We report the results in Table~\ref{tab:countermeasures_provable}}.

\section{Attack  Effectiveness  and  Generalization to Other   Tasks}
\label{sec:appendix_others}

\begin{figure}[h!]
    \centering
    \includegraphics[width=.7\linewidth]{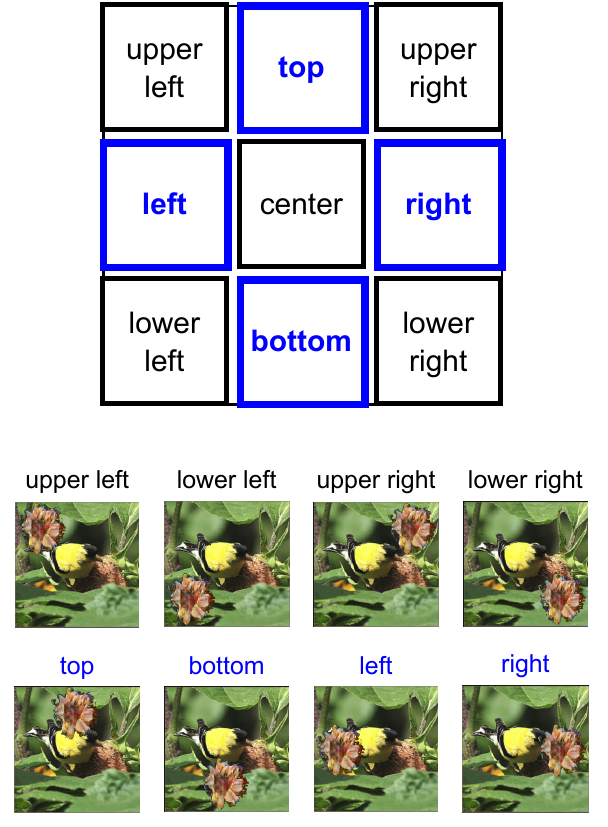}
    \caption{\hl{An illustration of trigger locations around the border.}}
    \label{fig:new_locations}
\end{figure}

\noindent\hl{\textbf{Effectiveness of Patch Locations}}. \hl{From Table~\ref{tab:locations}, we can see that with 8 locations around the border of the input images (exluding the center location), \tntss achieved the maximum ASR of 96.52\% (at upper-right corner) and the minimum ASR of 91.76\% (at left corner). Throughout all of 8 border locations, \tntss achieved a high mean of 94.4\% with a low variation of only 1.53\% showing the effectiveness of our \tntss even at border locations.}




\begin{table}[h!]
\centering
\caption{Illustrative examples of \tntss found in different visual classification tasks and their corresponding ASRs}
\label{tab:TnT-various}
\resizebox{\linewidth}{!}{%
\begin{tabular}{c c m{4cm} c}
\dtoprule
\textbf{Dataset}             & \textbf{ASR}  &  \textbf{Examples} & \textbf{Target} \\ \midrule
\texttt{CIFAR10}  

    
                         & 90.12\% & 

\begin{minipage}{\linewidth}
      \includegraphics[width=4cm, height=1cm]{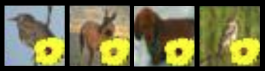}
    \end{minipage} &  \texttt{car}            \\ \midrule
\texttt{GTSRB}  

                         & 95.75\% & 

 \begin{minipage}{\linewidth}
      \includegraphics[width=4cm, height=1cm]{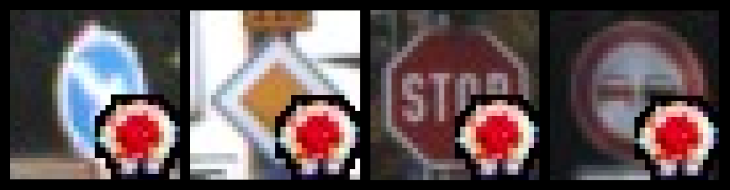}
    \end{minipage}                       
                        
& \texttt{untargeted} \\ \midrule

\texttt{PubFig} 


                          & 95.14\% & 

 \begin{minipage}{\linewidth}
      \includegraphics[width=4cm, height=1cm]{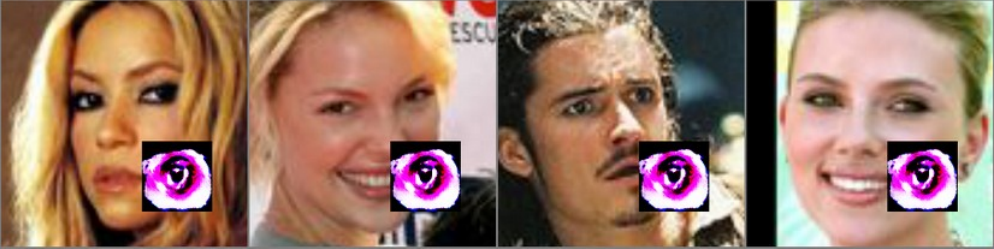}
    \end{minipage}                       
                        
& \texttt{untargeted} 

\\ \dbottomrule
\end{tabular}%
}
\end{table}


\section{Detailed Information On Datasets, Model Architectures and Training Configurations}
\label{sec:appendix_detailed_nets}

\captionsetup[table]{font=small}
\begin{table}[h!]
\centering
\small
\caption{\small{Networks used for the classification tasks}}
\begin{adjustbox}{width=0.9\columnwidth, center}
\begin{tabular}{ccccc} 
\dtoprule
\textbf{Task/Dataset} & \textbf{\makecell{\# of\\Labels}} &  \textbf{\makecell{\# of\\Training\\Images}} & \textbf{\makecell{\# of\\Testing\\Images}} & \textbf{\makecell{Model \\Architecture}} \\
\midrule
CIFAR10\cite{cifar10} & 10 & 50,000 & 10,000 & 6 Conv + 2 Dense \\ \midrule
GTSRB\cite{GTSRB} & 43 & 39,288 & 12,630 & 7 Conv + 2 Dense\\ \midrule
PubFig\cite{pubfig83} & 170 & 48,498 & 12,322 &  \makecell{13 Conv + 3 Dense\\(VGG-16)}\\ \midrule
ImageNet\cite{imagenet} & 1,000 & -\tablefootnote{In our work, we utilized pre-trained models on the ImageNet dataset from Pytorch~\cite{Pytorch}.} & 50,000 &  \makecell{WideResNet50~\cite{wideresnet}}\\
\dbottomrule
\end{tabular}
\label{table:networkstructure}
\end{adjustbox}
\end{table}

We describe the datasets used in our studies below.

\begin{itemize}
\item \textbf{Large Scale Visual Recognition} (\texttt{ImageNet}~\cite{imagenet}). ImageNet is a highly popular real-world dataset with a million high-resolution images of a large variety of objects used for training state-of-the-art deep perception models. The goal is to recognize visual scenes among 1,000 different classes. This is one of the most popular dataset in computer vision for benchmarks state-of-the-art models. In this task, we utilized state-of-the-art \textit{pre-trained} models available from Pytorch Deep Learning library~\cite{Pytorch}; notably, these models are used as base models by many Machine Learning practitioners for transfer learning to build systems for different visual tasks.  
\end{itemize}

Additionally, to demonstrate the generalization of our method, we also evaluate on 3 other visual classification tasks: i)~Scene Classification (\texttt{CIFAR10}); ii)~Traffic Sign Recognition (\texttt{GTSRB}); and iii) Face Recognition (\texttt{PubFig}). 


\begin{itemize}
        \item \textbf{Scene Classification} (\texttt{CIFAR10}~\cite{cifar10}). This is a widely used task and dataset with images of size $32\times32$ and we used a similar network to that implemented in the IEEE~S\&P~\cite{neuralcleanse} study.
    	\item \textbf{Traffic Sign Classification} (\texttt{GTSRB}~\cite{GTSRB}). German Traffic Sign Benchmark dataset is commonly used to evaluate vulnerabilities of DNNs as it is related to autonomous driving and safety concerns. The goal is to recognize 43 different traffic signs of size $32\times32$ to simulate a scenario in self-driving cars. 
    	The network we used follows the VGG~\cite{VGG} network structure. 
    	\item \textbf{Face Recognition} (\texttt{PubFig}~\cite{pubfig83}). Public Figures Face dataset is a large, real-world dataset with high-resolution images of large variations in pose, lighting, and expression. The goal is to recognize the faces of public figures. In this task, we leverage transfer learning from a publicly available pre-trained model based on a complex 16-layer VGG-Face model from the work of~\cite{Parkhi15} and fine-tune the last 6 layers.
\end{itemize}
\vspace{5mm}

\begin{table}[h!]
\centering
\caption{Model Architecture for VGGFace2}
\label{tab:vggface2_arch}
\begin{adjustbox}{width=.8\linewidth, center}

\begin{tabular}{ccccc}
\hline
Layer Type & \# of Channels & Filter Size & Stride & Activation \\ \hline
Conv       & 64             & 3           & 1      & ReLU       \\
Conv       & 64             & 3           & 1      & ReLU       \\
MaxPool    & 64             & 2           & 2      & -          \\
Conv       & 128            & 3           & 1      & ReLU       \\
Conv       & 128            & 3           & 1      & ReLU       \\
MaxPool    & 128            & 2           & 2      & -          \\
Conv       & 256            & 3           & 1      & ReLU       \\
Conv       & 256            & 3           & 1      & ReLU       \\
Conv       & 256            & 3           & 1      & ReLU       \\
MaxPool    & 256            & 2           & 2      & -          \\
Conv       & 512            & 3           & 1      & ReLU       \\
Conv       & 512            & 3           & 1      & ReLU       \\
Conv       & 512            & 3           & 1      & ReLU       \\
MaxPool    & 512            & 2           & 2      & -          \\
Conv       & 512            & 3           & 1      & ReLU       \\
Conv       & 512            & 3           & 1      & ReLU       \\
Conv       & 512            & 3           & 1      & ReLU       \\
MaxPool    & 512            & 2           & 2      & -          \\
FC         & 4096           & -           & -      & ReLU       \\
FC         & 4096           & -           & -      & ReLU       \\
FC         & 170            & -           & -      & Softmax    \\ \hline
\end{tabular}
\end{adjustbox}
\vspace{2mm}
\end{table}

\begin{table}[h!]
\centering
\caption{Model Architecture for GTSRB}
\label{tab:gtsrb_arch}
\begin{adjustbox}{width=.8\linewidth, center}

\begin{tabular}{ccccc}
\hline
Layer Type & \# of Channels & Filter Size & Stride & Activation \\ \hline
Conv       & 128            & 3           & 1      & ReLU       \\
Conv       & 128            & 3           & 1      & ReLU       \\
MaxPool    & 128            & 2           & 2      & -          \\
Conv       & 256            & 3           & 1      & ReLU       \\
Conv       & 256            & 3           & 1      & ReLU       \\
MaxPool    & 256            & 2           & 2      & -          \\
Conv       & 512            & 3           & 1      & ReLU       \\
Conv       & 512            & 3           & 1      & ReLU       \\
MaxPool    & 512            & 2           & 2      & -          \\
Conv       & 1024           & 3           & 1      & ReLU       \\
MaxPool    & 1024           & 2           & 2      & -          \\
FC         & 1024           & -           & -      & ReLU       \\
FC         & 10             & -           & -      & Softmax    \\ \hline
\end{tabular}
\end{adjustbox}
\vspace{2mm}
\end{table}



\begin{table}[h!]
\centering
\caption{Model Architecture for CIFAR-10. FC: fully connected layer.}
\label{tab:cifar10_arch}
\begin{adjustbox}{width=.8\linewidth, center}

\begin{tabular}{ccccc}
\hline
Layer Type & \# of Channels & Filter Size & Stride & Activation \\ \hline
Conv       & 128            & 3           & 1      & ReLU       \\
Conv       & 128            & 3           & 1      & ReLU       \\
MaxPool    & 128            & 2           & 2      & -          \\
Conv       & 256            & 3           & 1      & ReLU       \\
Conv       & 256            & 3           & 1      & ReLU       \\
MaxPool    & 256            & 2           & 2      & -          \\
Conv       & 512            & 3           & 1      & ReLU       \\
Conv       & 512            & 3           & 1      & ReLU       \\
MaxPool    & 512            & 2           & 2      & -          \\
FC         & 1024           & -           & -      & ReLU       \\
FC         & 10             & -           & -      & Softmax    \\ \hline
\end{tabular}
\end{adjustbox}
\end{table}

\begin{table*}[tp!]
\caption{Dataset and Training Configuration}
\label{tab:training_config}
\begin{adjustbox}{width=\linewidth, center}

\begin{tabular}{|c|c|c|c|c|c|}
\hline
\textbf{Task/Dataset} & \textbf{\# of Labels} & \textbf{Input Size} & \textbf{Training Set Size} & \textbf{Testing Set Size} & \textbf{Training Configurations}                                                                                                                                                                          \\ \hline
CIFAR-10              & 10 & $32\times32\times3$                    & 50,000                     & 10,000                   & \makecell{epochs=100, batch=32, \\optimizer=Adam, lr=0.001}                                                                                                                                         \\ \hline
GTSRB                 & 43  & $32\times32\times3$                  & 35,288                     & 12,630                    & \makecell{epochs=25, batch=32, \\optimizer=Adam, lr=0.001}                                                                                                                                          \\ \hline
PubFig                 & 83 & $224\times224\times3$                   & 11,070                     & 2,768                    & \makecell{epochs=30, batch=32, \\optimizer=Adam, lr=0.001. \\ The last 4 layers are fine-tuned during training.}                                                                                                                                         \\ \hline
ImageNet              & 1,000  & $224\times224\times3$                 & -                     & 50,000                  & We utilize pre-trained models available from Pytorch~\cite{Pytorch} for the task. \\ \hline
\end{tabular}
\end{adjustbox}
\vspace{5mm}
\end{table*}

\newpage
\begin{algorithm}[h!]
    \textbf{Inputs:} a batch of images $\{\bx^
		\text{(i)}\}_{i=1}^m$ with batch size $m$, source label $\{y^\text{(i)}_{\text{source}}\}_{i=1}^m$,
		target label $\{y^\text{(i)}_{\text{target}}\}_{i=1}^m$,
		model $p_M$, latent vector $\bz$, the hyper-parameter $\lambda$ to balance the loss, natural generator $G_\btheta$, and the thresholds to detect \tnts $\tau_\text{batch}$, $\tau_\text{val}$ for batch and validation set respectively.\;
	\textbf{Initialization:}\;
	\While {$ASR < \tau_\text{val}$}{
	    Sample a batch of images $\{\bx^\text{(i)}\}_{i=1}^m$\;
		Sample a latent variable $\bz \sim p(\bz)$\;
		$\bdelta = G_\btheta(\bz)$ \Comment{a flower patch}\;
		Generate the mask $\bmask$ based on $\bdelta$\;
		$\bdelta' \leftarrow \text{bgremoval}(\bdelta, \bmask)$\Comment{Background removal}\;
		\For {$i=1,...,m$}{
		    ${\bx'}^\text{(i)} = (1-\bmask)\odot \bx^\text{(i)}+\bmask\odot \bdelta'$\;
			$y^\text{(i)}_{\text{argmax}} = \arg\max_y p_M(y | {\bx'}^\text{(i)})$\;
			\If {$y^\text{(i)}_{\text{argmax}} = y_\text{target}^\text{(i)}$}{
			    $fool = fool + 1$\;
			}
		}
		$L=\ell(\{{\bx'}^\text{(i)}\}_{i=1}^m,\{y_{\text{target}}^\text{(i)}\}_{i=1}^m) - \lambda~ \ell(\{{\bx'}^\text{(i)}\}_{i=1}^m,\{y_{\text{source}}^\text{(i)}\}_{i=1}^m)$\;

        $\btheta \leftarrow \text{Adam}(\nabla_\btheta L,\btheta,\alpha,\beta_1,\beta_2)$\;
        \If {$fool > \tau_\text{batch} $}{
            Sample a latent variable $\bz \sim p(\bz)$\;
            $\bdelta = G_\btheta(\bz)$\;
            Test this $\bdelta$ for the whole validation set $\mathcal{X}_\text{val}$ to get $ASR$\;
            \If {$ASR \geq \tau_\text{val}$}{
                Complete update Generator, save the latest state as Adversarial Patch Generator $G_{\btheta'}$\;
            }
        }
	}
	\caption{Adversarial Patch Generator Process}
	\label{alg:adv-generator}
\end{algorithm}

\newpage

\section{\hl{Detailed Information On User Studies}}
\label{appd:user_study}

\hl{We adopted the Naturalistic Score and follow the procedure described in~\cite{hu2021naturalistic} to evaluate the human perception of our \tntss' naturalism using two study designs. As in~\cite{hu2021naturalistic}, naturalism was defined as the \textit{Naturalistic Score} of each patch calculated based on  the  percentage  of  participants’ votes for the patch. In contrast to~\cite{hu2021naturalistic}, employing 24 participants, we conduct a \textit{large} cohort user study with 250 participants for each of the following studies:}

\begin{figure}[h!]
    \centering
    \label{fig:user_studies}
    
    \begin{subfigure}{0.5\textwidth}
    \includegraphics[width=\textwidth]{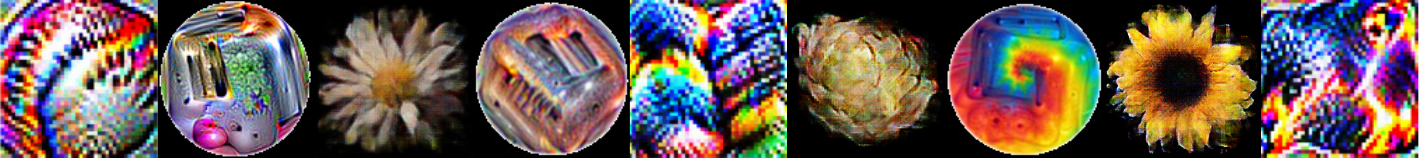}
    \caption{\hl{Study~1}}
    \label{fig:study1}
    \end{subfigure}
    ~
    \begin{subfigure}{0.35\textwidth}
    \includegraphics[width=\textwidth]{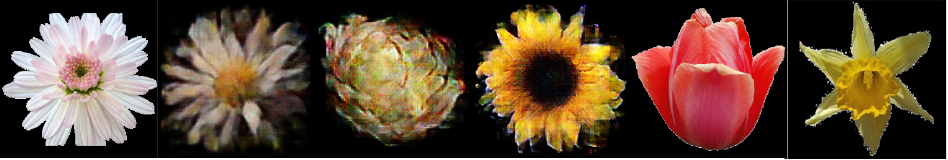}
    \caption{\hl{Study~2}}
    \label{fig:study2}
    \end{subfigure}
    \caption{\hl{An instance of the random ordering of patch images used in the two user studies, 250 participants participated in each study.}}
\end{figure}

\hl{\begin{itemize}
    \item \textbf{User study~1}. The aim of this study is to measure and compare the naturalistic score of our \tntss  with previous attacks. We used a set of 9 patch images; i)~3 patches generated by LAVAN~\cite{lavan}; ii) 3 generated by AdvPatch~\cite{adversarialpatch}; and iii)~3 \tntss. All the patches are placed in random order and shown to participants and the participants were asked to vote on each patch that looks natural to them (see Figure~\ref{fig:study1}). 
    \item \textbf{User study~2}. The aim of this study is to measure the absolute naturalistic score of our \tntss when compared with real flower images. We randomly placed 3 of our generated \tntss together with 3 real flower images collected from Google Images~\cite{googleimages} and asked participants to vote on the images that looks natural to them (see Figure~\ref{fig:study2}).
\end{itemize}}



\clearpage
\onecolumn

\section{Detailed Information On Black-box Attack}
\label{sec:appendix_blackbox}

Notably, to the best of our knowledge, we are the first to report qualitatively the untargeted black-box attack success rates where the patches---\tnts in our attack---do not occlude the main salient features of the images (Table~\ref{tab:blackbox}).

\begin{table}[h!]
\centering
\caption{Black-box attack, the transferability of \tnts from a model to other models on \texttt{ImageNet} dataset in an untargeted setting. ASRs are observed on 100 random images (network performance on the task is given in parenthesis).}
\label{tab:blackbox}
\resizebox{.7\linewidth}{!}{%
\begin{tabular}{rm{1cm}r|cccccc}
\dtoprule
&&&\multicolumn{6}{c}{\textbf{Target}} \\\cmidrule{4-9}
&\textbf{Example}& &
\textit{WideResNet50} &
  \textit{Inception V3} &
  \textit{ResNet18} &
  \textit{SqueezeNet 1.0} &
  \textit{VGG 16} &
  \textit{MnasNet} \\ \cmidrule{2-9}
&
\begin{minipage}{\linewidth}
      \includegraphics[width=1cm, height=1cm]{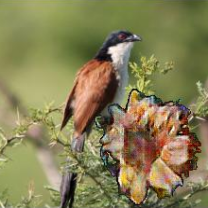}
    \end{minipage}
& \textit{WideResNet50} (Acc: 78.51\%)      & \textbf{97\%} & 77\% & 67\% & 77\% & 78\% & 63\% \\ \cmidrule{2-9}
&
\begin{minipage}{\linewidth}
      \includegraphics[width=1cm, height=1cm]{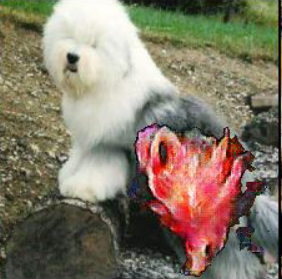}
    \end{minipage}
& \textit{Inception V3} (Acc: 77.45\%)      & 46\% & \textbf{91\%} & 51\% & 80\% & 66\% & 57\% \\ \cmidrule{2-9}
&
\begin{minipage}{\linewidth}
      \includegraphics[width=1cm, height=1cm]{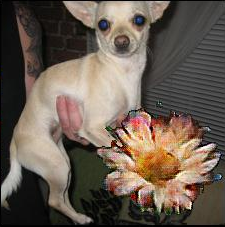}
    \end{minipage}
& \textit{ResNet18} (Acc: 69.76\%)           & 45\% & 47\% & \textbf{80\%}  & 64\% & 59\% & 54\% \\ \cmidrule{2-9}
\raisebox{\dimexpr \measureISpecification/2}[0pt][0pt]{\rotatebox[origin=c]{90}{\small \textbf{Source}}}
&
\begin{minipage}{\linewidth}
      \includegraphics[width=1cm, height=1cm]{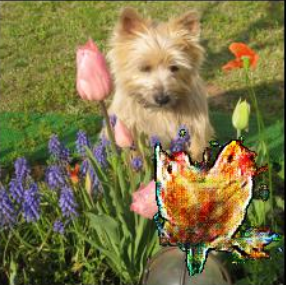}
    \end{minipage}
& \textit{SqueezeNet 1.0} (Acc: 58.1\%)     & 36\% & 42\% & 51\%  & \textbf{99\%} & 47\% & 48\% \\ \cmidrule{2-9}
&
\begin{minipage}{\linewidth}
      \includegraphics[width=1cm, height=1cm]{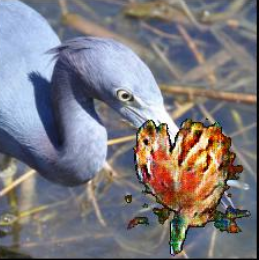}
    \end{minipage}
& \textit{VGG 16} (Acc: 71.59\%)    & 38\% & 49\% & 49\%  & 70\% & \textbf{91\%} & 49\% \\ \cmidrule{2-9}
&
\begin{minipage}{\linewidth}
      \includegraphics[width=1cm, height=1cm]{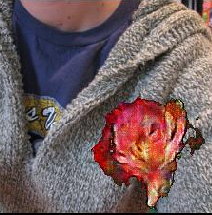}
    \end{minipage}
& \textit{MnasNet} (Acc: 73.51\%)            & 47\% & 63\% & 59\% & 74\% & 56\% & \textbf{87\%} \\ \dbottomrule
\end{tabular}%
}
\end{table}

\section{Illustrative Examples of Adversarial Patch Generator Outputs for Target Labels in LaVAN}
\label{sec:appendix_lavan}


\begin{figure}[h!]
    \centering
    \includegraphics[width=.7\linewidth]{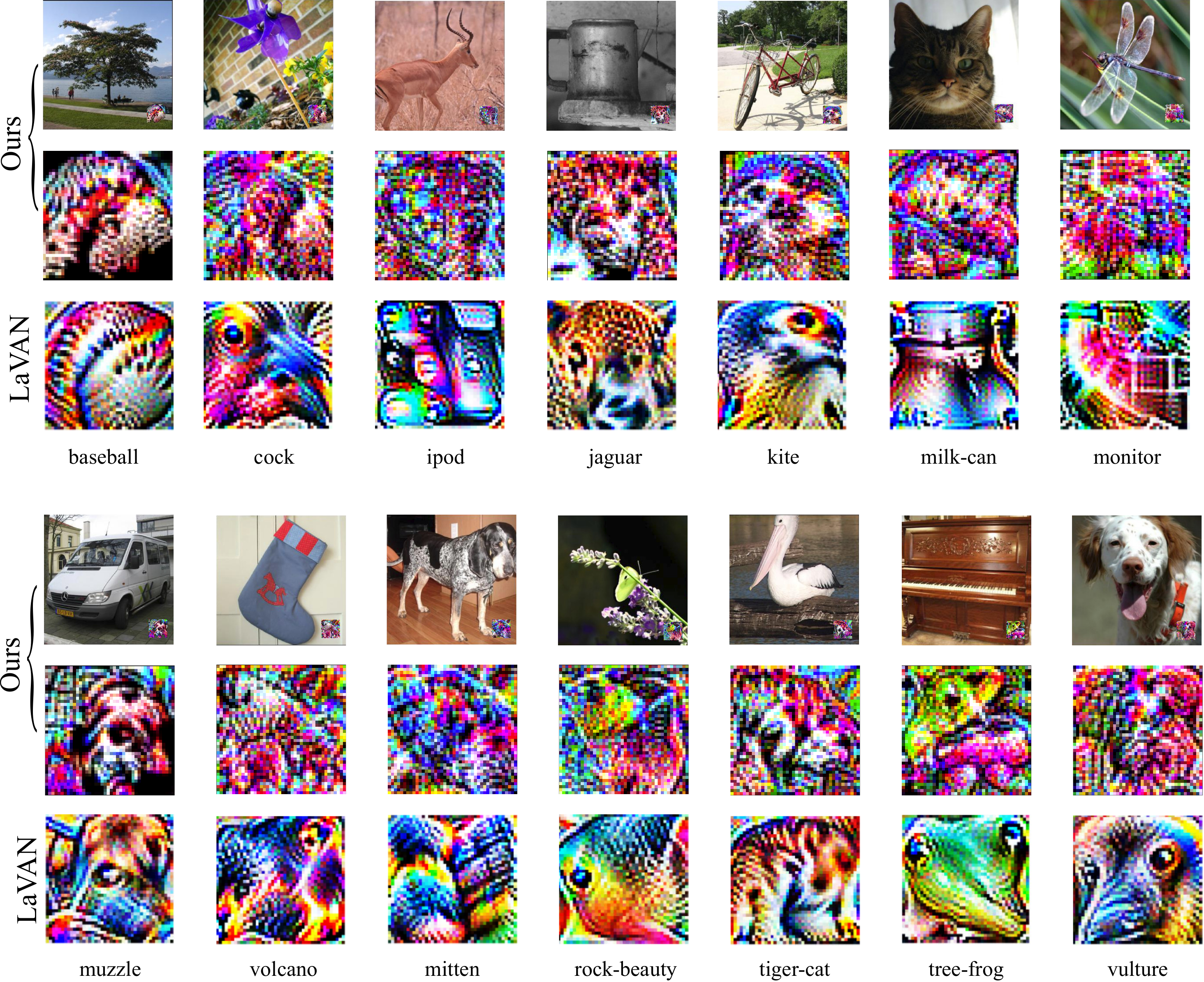}
    \caption{Generated patches from our Adversarial Patch Generator for the 14 different targeted labels in LaVAN. \textbf{The 1st row:} Our adversarial patches in the scene. \textbf{The 2nd row:} Our adversarial patches were rescaled to image size for display purposes. \textbf{The 3rd row:} adversarial patches from LaVAN for the same targeted label.}
    \label{fig:lavan-patches}
\end{figure}

\end{document}